%% file: camere_ready_final_arxiv.tex
\documentclass[runningheads]{llncs}

\usepackage{eccv}

\usepackage{eccvabbrv}

\usepackage{graphicx}
\usepackage{booktabs}

\usepackage[accsupp]{axessibility}  

\usepackage{hyperref}

\usepackage{orcidlink}
\usepackage{hyperref}

\usepackage{lipsum} 
\usepackage{soul}

\usepackage{algorithm}
\usepackage{algorithmic}

\usepackage{tabularx, booktabs}
\usepackage[export]{adjustbox}
\usepackage{multirow}
\usepackage{array}
\usepackage[normalem]{ulem}
\useunder{\uline}{\ul}{}
\usepackage{gensymb}
\usepackage{float}
\usepackage{caption}

\newcolumntype{P}[1]{>{\centering\arraybackslash}p{#1}}
\newcolumntype{Y}{>{\centering\arraybackslash}X}

\begin{document}

\title{Domain-Adaptive 2D Human Pose Estimation via Dual Teachers in Extremely Low-Light Conditions} 

\titlerunning{DA 2D-HPE via Dual Teachers in Extremely Low-Light Conditions}

\author{Yihao Ai\inst{1}\thanks{~Equal contribution.}\orcidlink{0009-0005-2336-4813}\and
Yifei Qi \inst{1}$^\star$\orcidlink{0009-0007-4197-947X}\and
Bo Wang\inst{2,3}\orcidlink{0000-0002-0127-2281}\and
Yu Cheng\inst{1}\orcidlink{0000-0002-9830-0081} \and \\
Xinchao Wang\inst{1}\orcidlink{0000-0003-0057-1404} \and
Robby T. Tan\inst{1}\orcidlink{0000-0001-7532-6919}}

\authorrunning{Y.~Ai et al.}

\institute{Department of Electrical and Computer Engineering, National University of Singapore \and 
Department of Computer and Information Science, University of Mississippi \and CtrsVision \\ \email{\{yihao, e0315880\}@u.nus.edu}, hawk.rsrch@gmail.com, e0321276@u.nus.edu, \{xinchao, robby.tan\}@nus.edu.sg}
\makeatletter
\let\@oldmaketitle\@maketitle
\renewcommand{\@maketitle}{\@oldmaketitle
  \input{figures/rep_samples}}
\makeatother

\maketitle

\begin{abstract}
    Existing 2D human pose estimation research predominantly concentrates on well-lit scenarios, with limited exploration of poor lighting conditions, which are a prevalent aspect of daily life. 
    Recent studies on low-light pose estimation require the use of paired well-lit and low-light images with ground truths for training, which are impractical due to the inherent challenges associated with annotation on low-light images.
    %
    To this end, we introduce a novel approach that eliminates the need for low-light ground truths.
    Our primary novelty lies in leveraging two complementary-teacher networks to generate more reliable pseudo labels, enabling our model achieves competitive performance on extremely low-light images without the need for training with low-light ground truths. Our framework consists of two stages.
    In the first stage, our model is trained on well-lit data with low-light augmentations.
    In the second stage, we propose a dual-teacher framework to utilize the unlabeled low-light data, where a center-based main teacher produces the pseudo labels for relatively visible cases, while a keypoints-based complementary teacher focuses on producing the pseudo labels for the missed persons of the main teacher. With the pseudo labels from both teachers, we propose a person-specific low-light augmentation to challenge a student model in training to outperform the teachers. 
    Experimental results on real low-light dataset (ExLPose-OCN) show, our method achieves \textbf{6.8\%} (\textbf{2.4} AP) improvement over the state-of-the-art (SOTA) method, despite no low-light ground-truth data is used in our approach, in contrast to the SOTA method. 
    %
    Our code is available at: \href{https://github.com/ayh015-dev/DA-LLPose}{DA-LLPose}. 
\end{abstract}

\section{Introduction}
\label{sec:intro}
Human 2D pose estimation is a fundamental task in computer vision and important for many downstream tasks like human activity recognition \cite{yan2018spatial}, virtual tryon \cite{dong2019towards}, motion capture \cite{desmarais2021review}, augmented reality \cite{weng2019photo}, or Metaverse \cite{jiang2022avatarposer}. 
The existing methods and benchmarks \cite{lin2014microsoft,zhang2019pose2seg,li2019crowdpose} primarily focus on well-illuminated scenarios, where human subjects exhibit both clear visibility and high recognizability in the majority of cases. 
However, poor lighting conditions such as low light or nighttime are a prevalent aspect of daily life, which signifies the importance of low-light human pose estimation research. 
Despite the significance of this research area, its exploration has been constrained by the lack of suitable low-light datasets or benchmarks.

A previous dataset for human pose estimation under poor lighting conditions primarily revolves around nighttime scenarios \cite{crescitelli2020rgb, crescitelli2020poison}, where human subjects generally remain recognizable due to the presence of light sources such as lamps or car lights. 
Thus, these nighttime datasets do not focus on extremely low-light conditions, which pose severe challenges to existing human pose estimation methods.
Recently, Lee et al. \cite{ExLPose_2023_CVPR} proposed a new benchmark for human pose estimation in extremely low-light conditions, which covers different levels of low-light conditions, and a low-light pose estimation method, which requires paired well-lit and low-light images with ground truths for training. 
However, the scarcity of paired images with ground truths in practical scenarios makes the application of this method impractical. 
To this end, we aim to develop an innovative domain adaptive method for low-light human pose estimation, utilizing well-lit ground-truth data only. 
Our goal is to achieve performance on par with the SOTA methods that rely on both low-light and well-lit ground truths.

To the best of our knowledge, we are the first to propose a solution to estimate human poses in extremely low-light conditions utilizing well-lit ground-truth data only. 
This is an extremely challenging task because the visibility of human subjects is poor and low-light images are corrupted due to severe low-light noise. The intensity and contrast are much lower in the low-light images (e.g., average pixel intensity of well-lit and low-light images are 90.5 and 2.0). 
Without paired well-lit and low-light images with ground-truth data, two types of existing methods could be used for low-light pose estimation: low-light image enhancement \cite{LLFlow, cai2023retinexformer} and domain adaptation \cite{AdvEnt, kim2022unified}. 
However, these two groups of methods have their own limitations. The former is difficult to restore the extremely low-light images faithfully and the latter still lacks a specific solution for bridging the gap between well-lit and low-light conditions. 

In response to these challenges, we introduce an innovative domain-adaptive dual-teacher framework for human pose estimation that facilitates knowledge transfer from well-lit to low-light domains. 
This innovation enables our model to exceed state-of-the-art (SOTA) methods by eliminating the requirement for training with low-light ground truths, which are often impractical to obtain. 
Ideally, cross-domain knowledge transfer is achievable using a single teacher, as demonstrated by existing methods \cite{tarvainen2017mean, kim2022unified, 2pcnet}. 
However, a single center-based teacher, designed to predict the entire human pose based on each person's center, may encounter difficulties in detecting individuals in low-light images due to inherent architectural limitations that focus solely on the detection of human centers.
Consequently, this approach may fail to identify individuals when their centers cannot be detected, resulting in a limited number of pseudo labels.
To generate a more comprehensive set of human pose predictions for use as pseudo labels, we propose the incorporation of a complementary teacher: a keypoint-based bottom-up human pose estimation model. This model is capable of predicting partial human poses by grouping detected keypoints, offering a complementary approach to the primary center-based teacher.

Our method comprises two stages:  pre-training and dual-teacher knowledge acquisition. During the first stage, we perform supervised training with labeled well-lit data alone. 
This initial step equips both teachers to recognize human poses in well-lit images and their corresponding low-light augmentations.
In the subsequent second stage, our focus shifts to harnessing knowledge directly from unlabeled low-light images.
To challenge the student model during the second stage, we introduce a novel augmentation technique called Person-specific Degradation Augmentation (PDA). 
PDA is selectively applied to images with pseudo labels generated by the teachers.
Given that low-light conditions pose a significant challenge in pose estimation, our degradation augmentations primarily darken the individuals, making them less distinguishable from their backgrounds to mimic realistic low-light scenarios. 
Such a simple, flexible and effective degradation neither necessitates the camera calibration as various physical-based augmentations did \cite{wei2020physics, wei2021physics}, nor relies on an additional network, which would cost extra computational resources \cite{Punnappurath_2022_CVPR}.
And it facilitates the student model's exploration of knowledge in the low-light domain, thereby encouraging it to surpass the performance of the teachers in the second stage.

\cref{fig:representative_result} shows three images from ExLPose-OCN \cite{ExLPose_2023_CVPR} and the human pose estimation results of the existing methods \cite{LLFlow,cai2023retinexformer,kim2022unified,AdvEnt} and our proposed method. 
%
Compared with the existing methods, our method can effectively estimate human poses in extremely low-light conditions. Specifically, our method is capable of handling low-contrast (last row) and partially visible (second row) persons. For the relatively visible persons (first row), our method can produce more accurate pose estimation. In summary, our major contributions are listed as follows. 

\begin{itemize}
  \item We investigate a challenging yet practical task, i.e., estimating human poses in extremely low-light conditions with well-lit ground-truth data only. To the best of our knowledge, we are the first to propose a solution in this challenging setting. 
  \item We introduce a novel domain-adaptive dual-teacher framework, utilizing both center-based and keypoint-based teachers to generate enriched pseudo labels for effective student model training, eliminating the need for low-light ground truths, which are often impractical to obtain. To our knowledge, this represents a significant advancement in low-light human pose estimation.
  \item Our proposed method outperforms the SOTA method by \textbf{6.8\%} (\textbf{2.4} AP) on the real low-light dataset, ExlPose-OCN. Notably, our method utilizes well-lit ground-truth data alone, while the SOTA method is trained with paired well-lit and low-light images with ground truths.
\end{itemize}

\section{Related Work} 
\noindent \textbf{2D Human Pose Estimation}
%
%
%
Two primary paradigms exist within 2D human pose estimation: top-down and bottom-up approaches. 
In general, top-down methods, benefiting from applying human detection first \cite{he2017mask,chen2018cascaded,xiao2018simple,tian2019directpose,sun2019deep}, tend to outperform their bottom-up counterparts \cite{cao2019openpose,newell2017associative,kocabas2018multiposenet,cheng2020higherhrnet}. 
Early approaches of bottom-up estimation \cite{cao2019openpose, newell2017associative} are keypoint-based, detecting keypoints for each individual in an image before grouping them to construct individual human poses. 
Recently,  Geng et al. propose a center-based bottom-up method \cite{DEKR2021}, which predicts human centers instead of grouping individual keypoints, and regresses offsets between a human center and the keypoints for each individual. 
This method sets a new trend among the bottom-up methods \cite{LOGOCAP, wang2022contextual,wang2021robust}.
However, existing 2D pose estimation methods focus on well-lit conditions, primarily due to the prevalence of well-lit images within the benchmark datasets \cite{lin2014microsoft,zhang2019pose2seg,li2019crowdpose}. 
Additionally, since existing methods are fully supervised, they demonstrate poor performance when applied directly to low-light images after being trained on well-lit data.

\noindent \textbf{Domain Adaptive Pose Estimation}
Domain adaptive pose estimation is potentially useful for low-light conditions. 
A number of domain adaptation methods are proposed in human pose \cite{kim2022unified, raychaudhuri2023prior, peng2023source}, animal pose \cite{mu2020learning, cao2019cross, li2021synthetic} and hand pose estimations \cite{jiang2021regressive, han2022learning, jin2022multibranch}.  
%
%
%
However, these methods are not designed for human pose estimation in extremely low-light conditions. 
%
%
Source-free domain adaptive human pose methods are developed \cite{raychaudhuri2023prior,peng2023source} to transfer knowledge from synthetic to real data in the absence of the source data. This is in opposition to low-light scenarios, where well-lit (source) domain data is typically available.
%
Kim et al. propose a unified framework for domain adaptive human pose estimation \cite{kim2022unified}, where the student is challenged to learn the target-domain knowledge by applying affine transform-based augmentations. However, this augmentation is not tailored for the severe lighting conditions. 
%
%

\noindent \textbf{Low-light Human Pose Estimation} Apart from the domain adaptation-based methods, fully-supervised approaches are developed as well. Crescitelli et al. propose a nighttime human pose estimation method that combines images from both RGB and infrared cameras \cite{crescitelli2020rgb,crescitelli2020poison}. 
However, the reliance on infrared cameras restricts the practical applicability of this method. 
Recently, Lee et al. \cite{ExLPose_2023_CVPR} propose a top-down human pose estimation approach for extremely low-light conditions. 
Nevertheless, this method depends on the use of paired well-lit and low-light images with ground-truth data during training, which are impractical for real-world applications due to the scarcity of the paired data and the inherent challenges associated with annotating low-light images. 
%
In the context of semi-supervised learning for human pose estimation, the utilization of multiple teachers, especially in dual-teacher configurations \cite{xie2021empirical, huang2023semi}, is related to our work. However, semi-supervised learning typically leverages both labeled and unlabeled data under the assumption that they come from similar distributions. 
This assumption differs significantly from the low-light setting, underscoring the necessity for our innovative domain-adaptive dual-teacher framework.

\noindent \textbf{Low-light Image Enhancement}
Low-light image enhancement might be used to enhance low-light images prior to human pose estimation. 
Early approaches primarily rely on gamma correction \cite{rahman2016adaptive} or histogram equalization \cite{cheng2004simple, celik2011contextual}. 
More advanced methods restore the low-light images based on the Retinex theory \cite{li2018structure, wang2013naturalness}. 
Unfortunately, these methods tend to introduce additional noise with color distortion. 
%
In the deep learning era, a number of convolutional neural network (CNN)-based \cite{moran2020deeplpf, sharma2021nighttime, wang2019underexposed} and GAN-based \cite{jin2022unsupervised, huang2017arbitrary} methods are developed. 
%
%
Recently, Wang et al. introduce an invertible network to capture the distribution of the well-lit images \cite{LLFlow}. 
%
Cai et al. propose a transformer-based approach to restore the hidden corruption in low-light images \cite{cai2023retinexformer}. 
However, these image enhancement methods still face difficulties in faithfully restoring extremely low-light images. 
Thus, even with enhanced images of these methods, low-light pose estimation is still challenging.

\begin{figure*}[t]
	\centering
	\includegraphics[width=\linewidth]{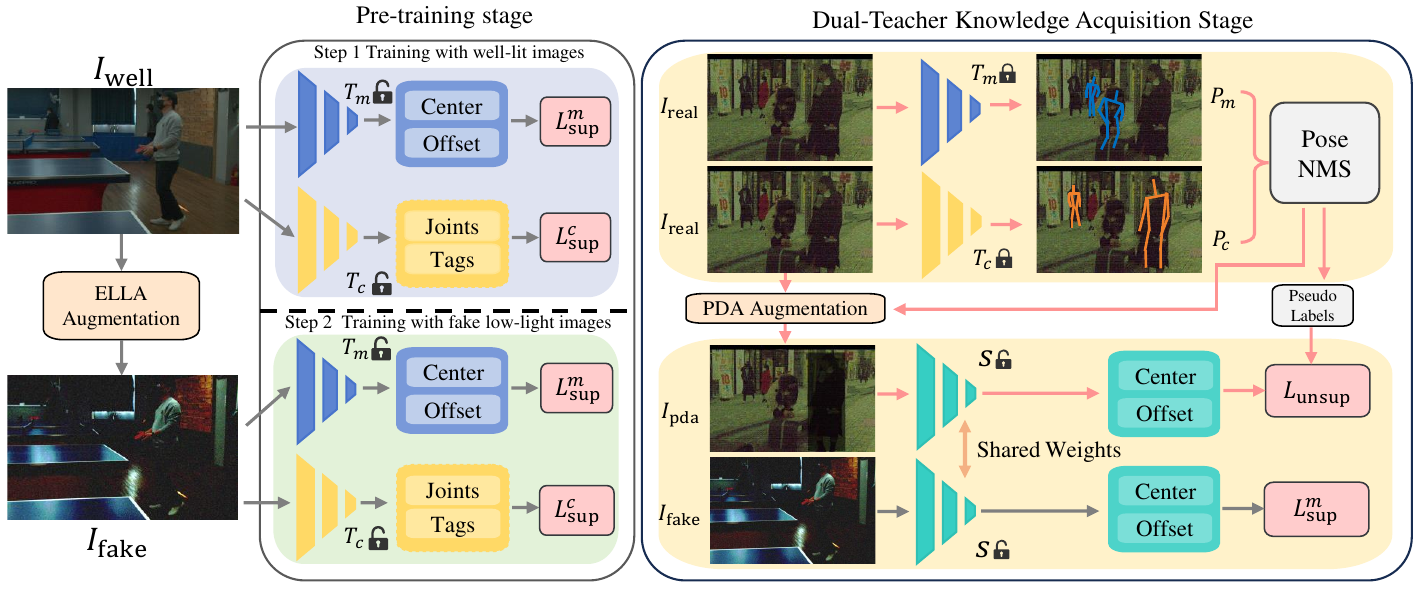}
	\caption{The overview of our method. The grey uni-directional arrows indicate the supervised workflow in the pre-training stage. The grey and pink arrow indicates the workflow of the supervised and unsupervised losses in the knowledge acquisition stage. The low-light images, $I_{\rm fake}$, $I_{\rm real}$ and $I_{\rm pda}$ are all brightened for improved visibility. NMS stands for non-maximum suppression.}
	\label{fig:overview}
\end{figure*}

\section{Proposed Method}

Given an image $I\in{\mathbb{R}^{3\times{h}\times{w}}}$ which contains persons in the low-light conditions, 2D human pose estimation is to obtain the human skeletons from an image by locating $K$ types of human keypoints for each person. An overview of our proposed framework is shown in \cref{fig:overview}, where we present a novel domain-adaptive dual-teacher framework, utilizing both center-based and keypoint-based teachers to generate enriched pseudo labels for effective student model training, as shown in the yellow boxes. 
%
\subsection{Main Teacher and Complementary Teacher}
\paragraph{Main Teacher} We build our main teacher based on DKER \cite{DEKR2021}, where a person's pose is represented as the center and offset in addition to the heatmap for each keypoint. In particular, the center location of the $i$-th person is $\textbf{c}^i = \frac{1}{K}\sum_{k=1}^K\textbf{p}^i_{k}$ and the offsets from the human center to each keypoints are $\textbf{o}^i=\{\textbf{p}^i_1 - \textbf{c}^i, \textbf{p}^i_2 - \textbf{c}^i, ..., \textbf{p}^i_k - \textbf{c}^i\}$, where $\textbf{p}^i_{k} \in{\mathbb{R}^2}$ indicates the 2D location of the $k$-th keypoints for the $i$-th person. The model is supervised by both heatmap and offset losses:
\begin{equation}
	L_{\rm sup}^{m} = L_H^{m} + \lambda_m L_O^{m}, 
	\label{equ:lsup_main}
\end{equation}
where $L_H^{m}$ denotes a MSE loss of the main teacher's heatmaps, and $L_O^{m}$ is a smooth L1 loss of the offset map.

\paragraph{Complementary Teacher} The main drawback of the center-based representation is that the pose estimation fully relies on the center's predictions. If the center of a person is failed to be detected, the whole person is missed, especially for the low-contrast and partially visible persons in the low-light conditions. 
Therefore, we build a complementary teacher to predict human poses for partially visible persons following HigherHRNet-style design \cite{cheng2020higherhrnet}. 
Specifically, the network is trained to estimate the keypoints' heatmap $\hat{H}_{c}$ and the corresponding tag map $\hat{T}_c$. A person's pose is grouped based on all the identity information via Hungarian algorithm \cite{kuhn1955hungarian}. The method is supervised by following losses:

\begin{equation}
	L_{\rm sup}^{c} = L_H^{c} + \lambda_c L_{\rm tag}^{c}, 
	\label{equ:lsup_comp}
\end{equation}
where, $L_H^{c}$ is a MSE loss of the complementary teacher's heatmaps, and $L_{\rm tag}^{c}$ consists of push and pull losses of the tag maps. More details are provided in the supplementary.

\subsection{Pre-Training Stage}
Firstly, the two teachers are trained on the well-lit dataset \cite{ExLPose_2023_CVPR} to gain the common knowledge of the human bodies. They are supervised by the losses defined in \cref{equ:lsup_main} and \cref{equ:lsup_comp} respectively.
After the supervised training with the labeled well-lit data, teachers can predict human poses on well-lit images, but they cannot process low-light images due to the substantial domain gap between well-lit and low-light conditions. 
As our framework relies solely on well-lit ground-truth data, we propose Extreme Low-Light Augmentation (ELLA) to augment low-light images from well-lit ones. This process involves incorporating a set of low-light characteristics into the well-lit images.

\begin{figure}[t]
    \centering

    \begin{subfigure}{0.159\textwidth}
        \centering
        \includegraphics[width=\linewidth]{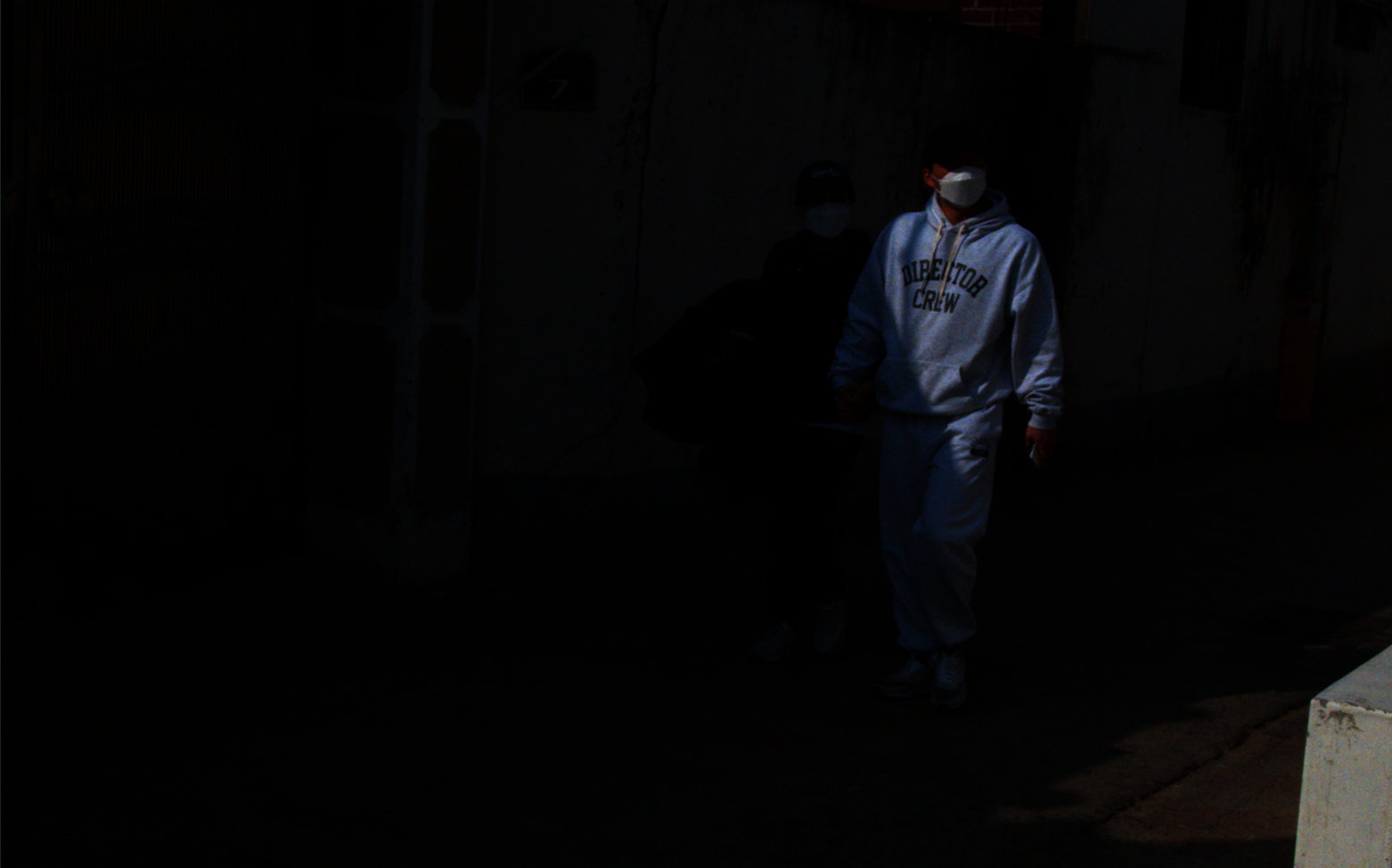}
        \caption{Gamma}
    \end{subfigure}
    \hfill
    \begin{subfigure}{0.159\textwidth}
        \centering
        \includegraphics[width=\linewidth]{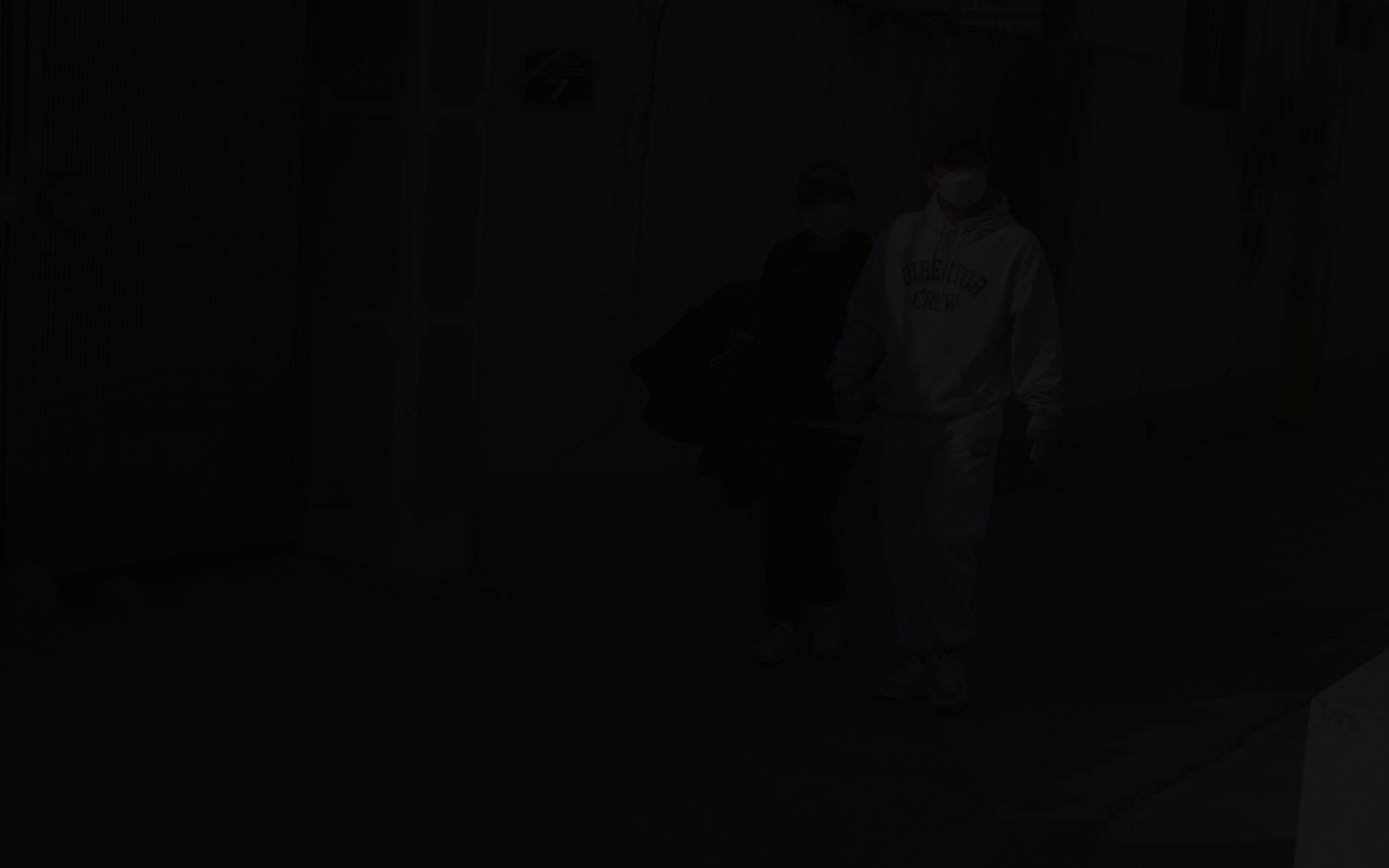}
        \caption{Brightness}
    \end{subfigure}
    \hfill
    \begin{subfigure}{0.159\textwidth}
        \centering
        \includegraphics[width=\linewidth]{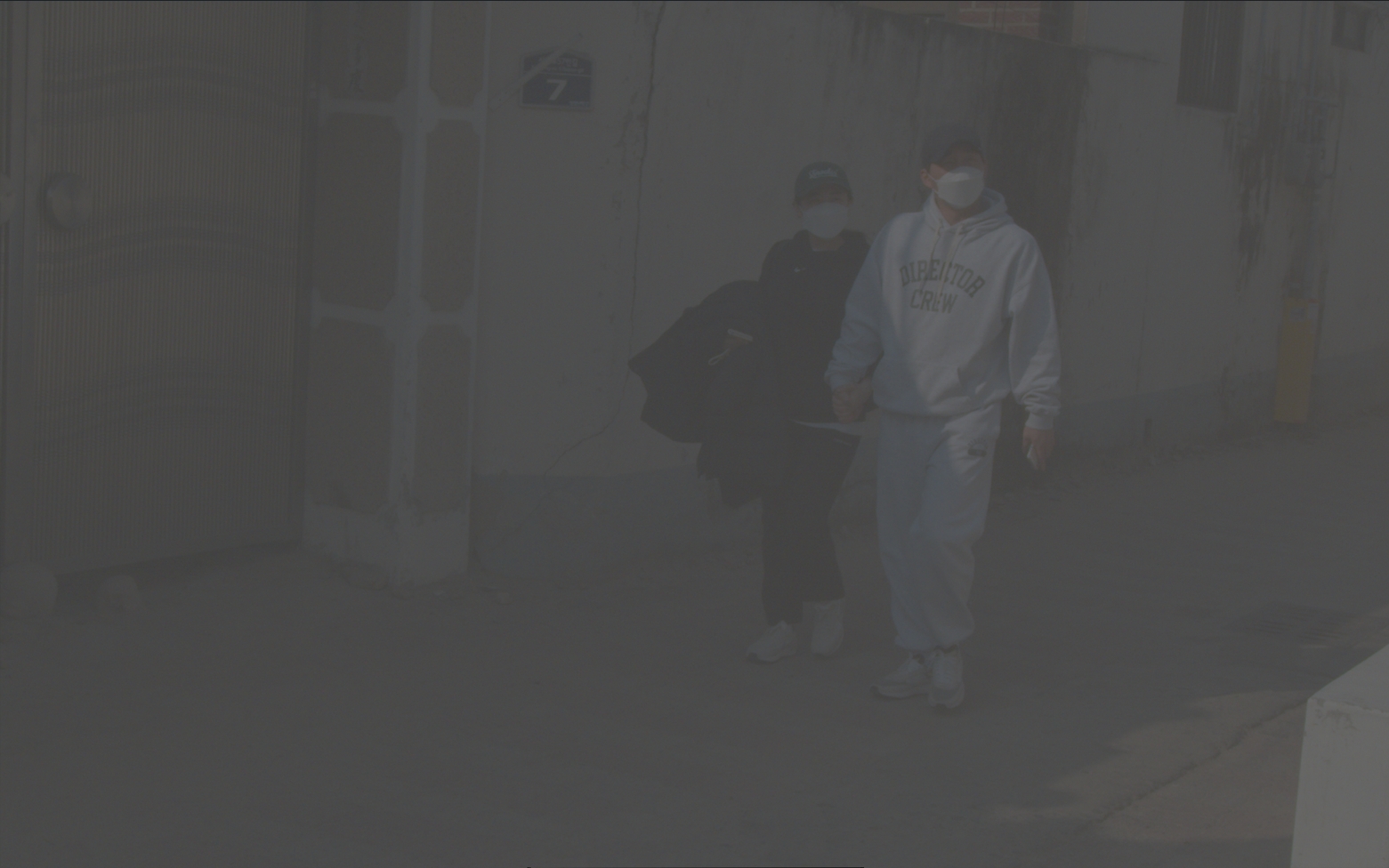}
        \caption{Contrast}
    \end{subfigure}
    \begin{subfigure}{0.159\textwidth}
        \centering
        \includegraphics[width=\linewidth]{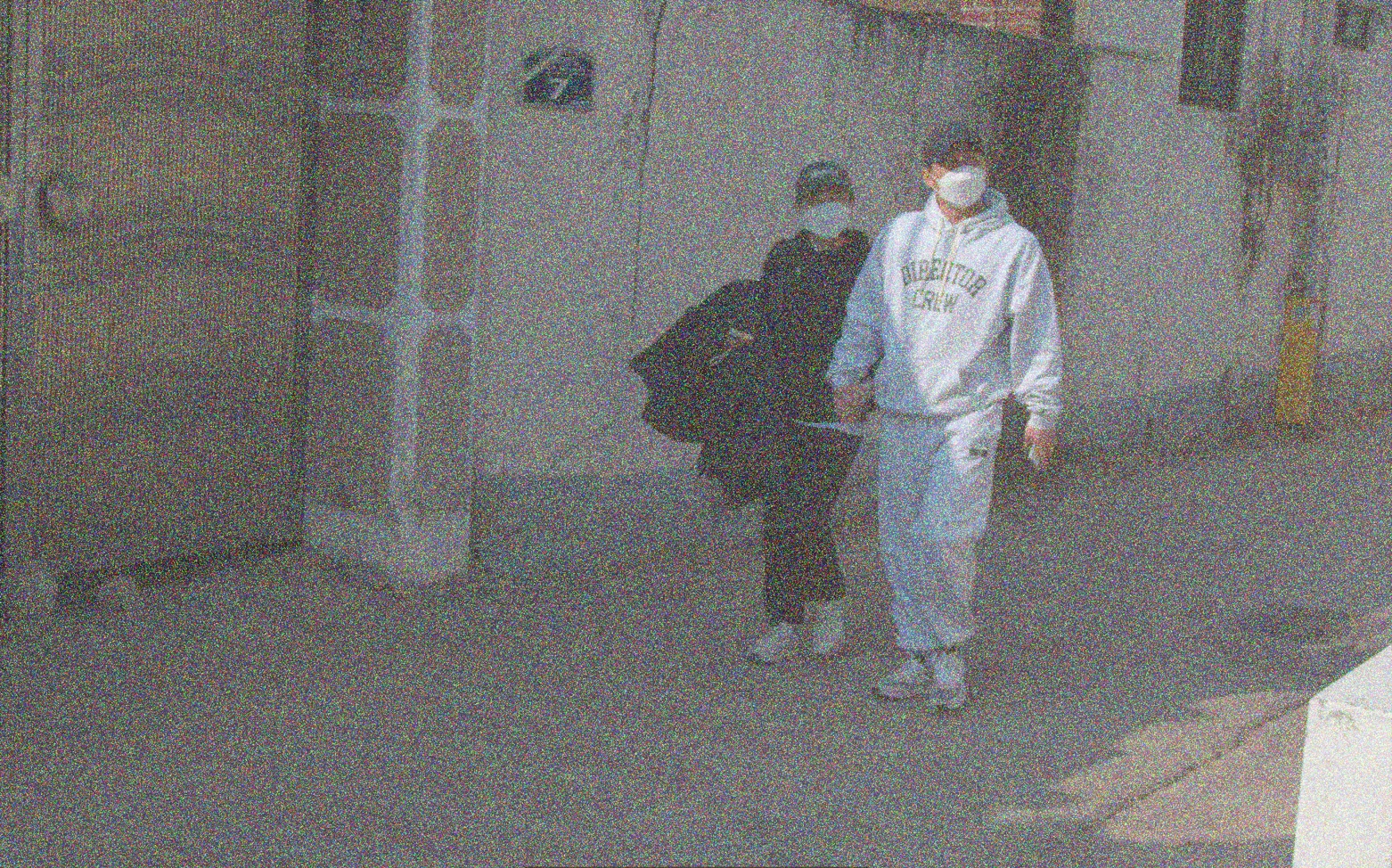}
        \caption{Noise}
    \end{subfigure}
    \hfill
    \begin{subfigure}{0.159\textwidth}
        \centering
        \includegraphics[width=\linewidth]{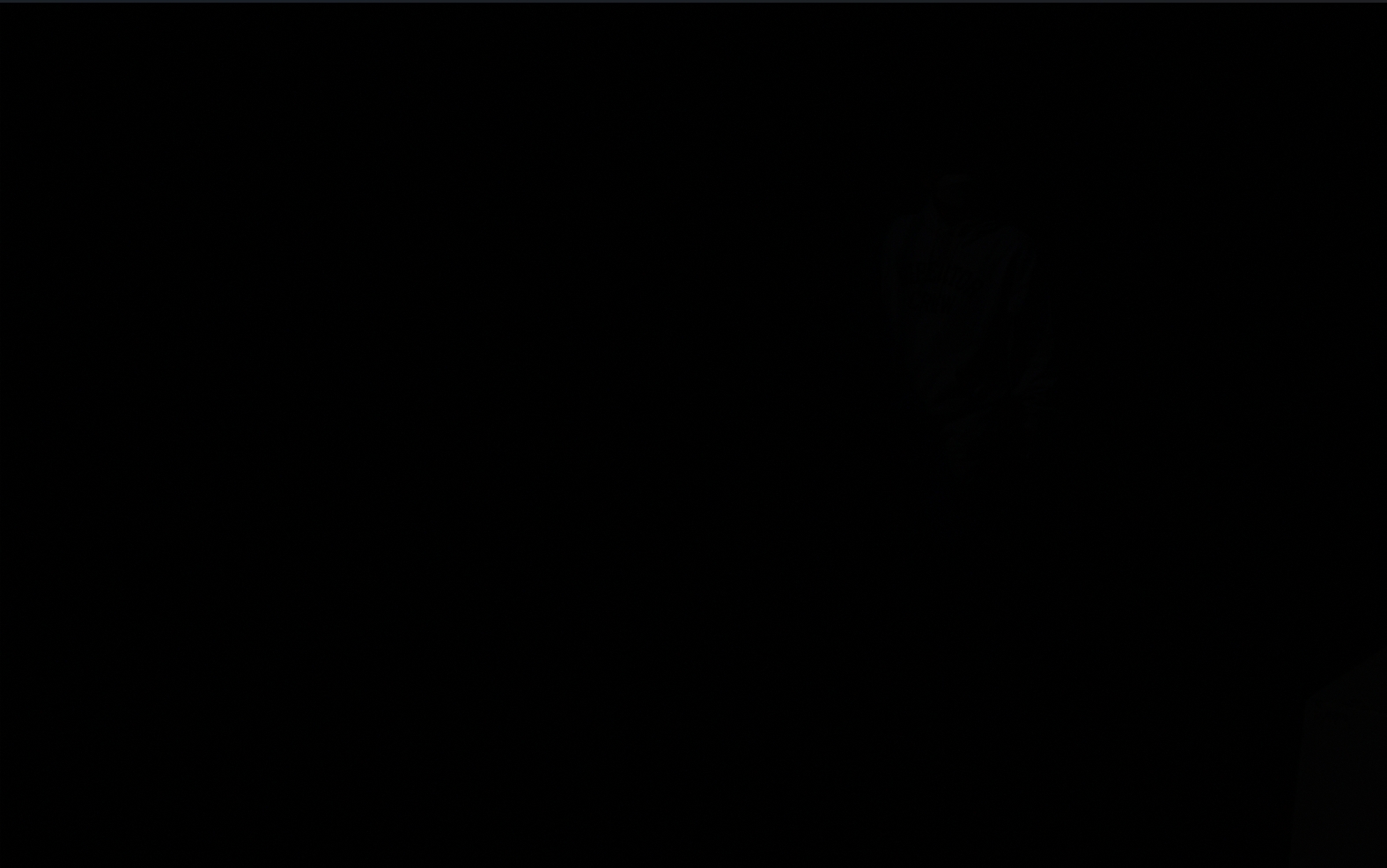}
        \caption{\rm All Aug.}
    \end{subfigure}
    \hfill
    \begin{subfigure}{0.159\textwidth}
        \centering
        \includegraphics[width=\linewidth]{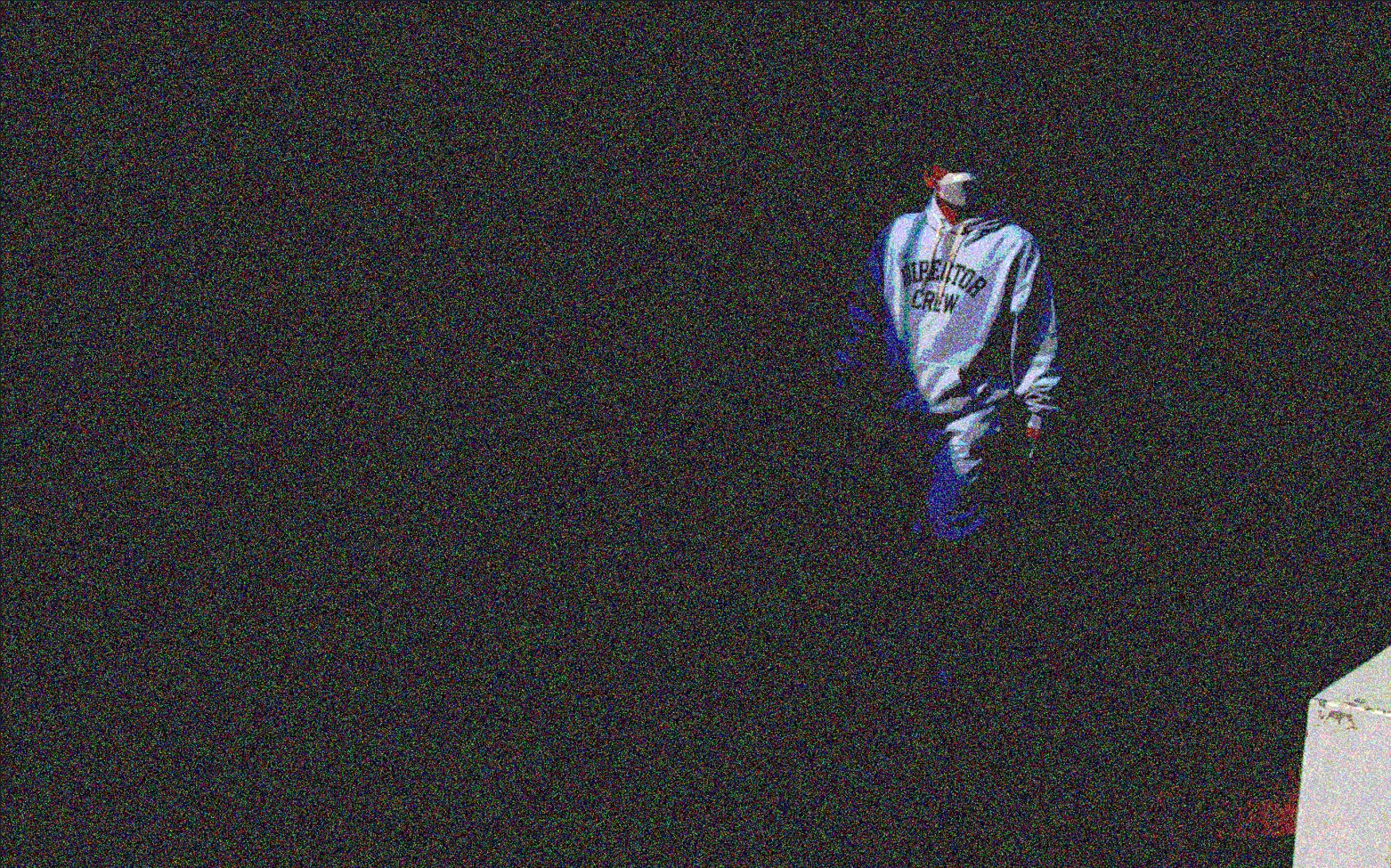}
        \caption{Bri. All}
    \end{subfigure}

    \caption{Examples of our ELLA. (a)-(d) are the results of the individual augmentation in ELLA. (e) is the image with all the augmentations, and Aug. stands for augmentations. (e) is brightened up for improved visibility as shown in (f), where Bri. All denotes brightened images with all the augmentations.}
    \label{fig:ELLA}
\end{figure}

The distinguishing characteristics of low-light images include extreme darkness, high levels of noise, and low contrast. 
To simulate these low-light specific features, ELLA applies a series of random augmentations sequentially to well-lit images, including gamma correction, brightness adjustment, contrast reduction, and Gaussian noise. 
The first two augmentations aim to intensify the darkness of the images, while contrast reduction is employed to decrease image contrast, and Gaussian noise is added to introduce noise into the images. 
It is important to note that in ELLA, we do not explicitly simulate light sources. This is because our primary focus is on low-light conditions, where no light source is typically available.
An example of the augmented images using ELLA is shown in \cref{fig:ELLA}. 
The following equations, \cref{equ:gamma}, \cref{equ:bright}, \cref{equ:contrast}, and \cref{equ:Gauss}, mathematically represent the four ELLA augmentations:

\begin{center}
\begin{minipage}[c]{0.48\linewidth}
\begin{align}
     I_{\rm out} &= 255 \cdot \left(I_{in} / 255\right)^\gamma \label{equ:gamma}, \\
     I_{\rm out} &= bI_{\rm in} \label{equ:bright},
\end{align}
\end{minipage}
\begin{minipage}[c]{0.48\linewidth}
\begin{align}
     I_{\rm out} &= cI_{\rm in} + (1-c)I_{\rm grey} \label{equ:contrast},\\
     I_{\rm out} &= I_{\rm in} + {\rm Gaussian}(0, var) \label{equ:Gauss},
\end{align}
\end{minipage}
\end{center}

where $I_{\rm in}$ is the input image, $I_{\rm grey}$ is the greyscale image of $I_{\rm in}$, and $I_{\rm out}$ stands for the output image. $\gamma \in [2, 5]$, $b \in [0.01, 0.05]$, $c \in [0.2, 1.0]$ and $var \in [0, 40]$ are the parameters uniformly sampled from their respective ranges.

The augmented well-lit images are named fake low-light images, denoted as $I_{\rm fake}$. 
The augmentation is added randomly so that both teachers are trained with a mixture of well-lit images and fake low-light images and supervised by the losses in \cref{equ:lsup_main} and \cref{equ:lsup_comp} respectively (each of the ELLA augmentations has a 0.5 probability to be added). 
It is important to use both well-lit images and fake low-light images in the training because it is difficult for the model to transfer the knowledge to low-light domain if all inputs suddenly change to the fake low-light images.

\subsection{Dual-Teacher Knowledge Acquisition Stage}
After the pre-training stage, the teachers possess the preliminary capability to perform human pose estimation for extremely low-light images.
In this stage, our primary objective is to enhance the student model's ability for detecting and localizing human keypoints by leveraging real low-light images and guiding the student to surpass both teachers. 
In particular, given a non-paired fake low-light image $I_{\rm fake}$, a real low-light image $I_{\rm real}$, and the pre-trained two teachers $T_m$, $T_c$, we train the student $S$, who shares the same architecture of the main teacher at the beginning, by a supervised loss and an unsupervised loss. 
%
The workflow of the supervised and unsupervised losses is illustrated in \cref{fig:overview} with grey and pink arrows respectively. 

\paragraph{Supervised Loss} 
The supervised loss is a continuation of the pre-training stage. To strengthen the student's knowledge on the low-light domain, a larger proportion of fake low-light images is expected to be added in this stage. 
Ideally, all well-lit images that are used in the student's training should be augmented in the low-light fashion by ELLA. However, this would trigger the knowledge forgetting problem \cite{lin2022prototype, wang2022continual} which is an inherent problem in domain adaptation methods. 
%
To strengthen the learning in the target domain while keeping the source knowledge, we make a few adjustment to the ELLA as follows.

\paragraph{Adjust ELLA} On the one hand, we set ELLA to do gamma correction and brightness adjustment as \cref{equ:bright} and \cref{equ:gamma} to every input well-lit images. This aims to augment more samples in the low-light domain. 
On the other hand, to avoid the forgetting problem, we randomly crop a few image patches in an augmented image and restore it to the well-lit domain. The cropping probability for an image is 0.15 to guarantee the majority of the inputs are still fully-augmented images. 
The supervised loss is computed by the following equations (shown in \cref{fig:overview} with grey arrows). First, well-lit images $I_{\rm well}$ are augmented to fake low-light images $I_{\rm fake}$ with adjusted ELLA as:
\begin{equation}
   I_{\rm fake} = {\rm AdjustELLA}(I_{\rm well}).
   \label{equ:adjust_ella}
\end{equation}
Second, student $S$ produces the prediction for $I_{\rm fake}$:
\begin{equation}
   \hat{H}_f, \hat{O}_f = S(I_{\rm fake}),
   \label{equ:stu_fake_output}
\end{equation}
where $\hat{H}$ and $\hat{O}$ refer to the heatmaps and offset maps, which are used in the calculation of the supervised loss following \cref{equ:lsup_comp}.

\paragraph{Unsupervised Loss}
Apart from the supervised loss, we teach the student with the pseudo labels selected and fused by the predictions from both teachers. 
As shown in the \cref{fig:overview}, the two teachers firstly generate their own pose predictions $P_m$ and $P_c$ by $P_m = T_m(I_{\rm real})$ and $P_c = T_c(I_{\rm real})$.
Since the two teachers score the poses in different manner (the main teacher directly uses the center's activation, $C_m$, while the complementary teacher averages the heatmap activation of all keypoints as the confidence score, $C_c$), we use different thresholds $s_m$ and $s_c$ to select out the high-confidence poses. 
Then the poses from both teachers are concatenated by and applied non-maximum suppression (NMS) \cite{hosang2017learning} to rule out the redundant poses predictions to obtain the pseudo labels $P_{\rm all}$ defined as:
\begin{equation}
   \label{equ:PoseNMS}
   P_{\rm all} = {\rm NMS}(P_m[C_m > s_m] \oplus P_c[C_c > s_c]),
\end{equation}
where $\oplus$ indicates concatenation operation.

\paragraph{Person-specific Degradation Augmentation}
The student needs to gain low-light domain-specific knowledge in order to outperform the teachers in challenging scenarios under low illumination.
%
%
For example, the right person with black shirt in the last row of \cref{fig:representative_result} is largely indistinguishable from the dark background, which demonstrates that low contrast is a major challenge for low-light human pose estimation. 
Therefore, the network has to learn the specific human structure in a texture-less condition. To this end, we propose a Person-specific Degradation Augmentation (PDA) to realistically mimic the extremely low-light conditions at individual-level. 
%
%
Specifically, after getting the selected poses $P_{\rm all}$ from the teachers, a set of bounding boxes is generated to crop the individuals. 
The cropped image patches are augmented based on {\cref{equ:bright}}, which are then used to replace the original ones from input $I_{\rm real}$ to generate output $I_{\rm pda}$. 

The last step of the unsupervised loss computation is to generate the pseudo labels based on $P_{\rm all}$ and to supervise the output of the student:
\begin{equation}
 \hat{H}_r, \hat{O}_r = S({\rm PDA}(I_{\rm real}, P_{\rm all})).
 \label{equ:stu_real_output}
\end{equation}
The unsupervised loss $L_{\rm unsup}^{m} = L_H^{m} + \lambda_m L_O^{m}$ is similar to \cref{equ:lsup_main}. 
Combining both supervised and unsupervised losses, the loss of the student $S$ is defined as follows: 
\begin{equation}
 L{s} = \lambda_{\rm sup} L_{\rm sup}^{m} + \lambda_{\rm unsup} L_{\rm unsup} .
 \label{equ:stu_loss}
\end{equation}


\section{Experiments}
\subsection{Implementations and Dataset}
\paragraph{Dataset}
ExLPose datasets \cite{ExLPose_2023_CVPR} are used in validating our method, which is a new dataset that is specifically designed for evaluating 2D human pose estimation under extremely low-light conditions. 
The training set of ExLPose provides $2,065$ pairs of well-lit and low-light images and the ground-truth definition follows the CrowdPose \cite{li2019crowdpose} format. 
These images were taken from 251 indoor and outdoor scenes in the daytime and the paired low-light images are artificially produced by using a dual-camera system with different settings to ensure a wide coverage of diverse low-light conditions. 

There are two different testing sets in ExLPose \cite{ExLPose_2023_CVPR}. One is ExLPose-OCN , in which 360 extremely low-light images are captured at night with two different cameras, A7M3 and RICOH3. Another one is ExLPose-test, which consists of 491 man-made low-light images captured and produced similarly as those in the training set. Low-light All (LL-A) is used to denote the whole ExLPose-test, which is further divided into three subsets according to their difficulty level including Low-Light Normal (LL-N), Hard (LL-H), and Extreme (LL-E).
%

Among the two testing sets, ExLPose-OCN is captured in real low-light environments, while the low-light images in the ExLPose-test are obtained by artificially reducing the amount of light by 100 times \cite{ExLPose_2023_CVPR}. 
In addition, ExLPose-OCN is a test-only dataset, which is used to evaluate the generalization capability of a model trained on ExLPose-test.
Therefore, the evaluation on ExLPose-OCN provides a more objective measurement of the performance of low-light human pose estimation. 
In our method, we only use the unpaired labeled well-lit images and unlabeled low-light images in the training set to train our models. Low-light ground-truth data is never used in training our model. 


\begin{table*}[t]
     	\footnotesize
     	\centering
      \caption[Evaluation on ExLPose-OCN dataset]{Evaluation on ExLPose-OCN with top-down  baselines (lower part of the table) and the bottom-up baselines (upper part of the table) tested in our paper. Our method is a bottom-up method. The reported top-down baselines are all fully-supervised methods. LL indicates low-light labels. WL indicates well-lit labels. The best is \textbf{bold}. The second best is {\ul underlined}. }
     	\resizebox{0.7\columnwidth}{!}{
            \begin{tabularx}{0.8\textwidth}{l| *{4}{Y|} Y}
     		\hline
     		\hline
             
     		\multirow{2}{*}{\textbf{Methods}} & \multicolumn{2}{c|}{\textbf{Training Labels}} & \multicolumn{3}{c}{\textbf{AP$^{\uparrow}$@0.5:0.95}} \\
     		\cline{2-6}
       \rule{0pt}{2.6ex}
     		 & LL & WL & A7M3 & RICOH3 & Avg. \\
     		\cline{1-6}
     	    Base-low \cite{DEKR2021} &\checkmark & & 27.1 & 15.9 & 21.5 \\
     	    Base-well \cite{DEKR2021} & &\checkmark & 5.3 & 7.4 & 6.3 \\
         RFormer \cite{cai2023retinexformer} & &\checkmark & 18.9 & 17.7 & 18.3 \\
     	LLFlow \cite{LLFlow} & &\checkmark & 23.7 & 19.1 & 21.4 \\
     UDA-HE \cite{kim2022unified} & &\checkmark & 6.5  & 7.2 & 6.9 \\
     	    AdvEnt \cite{AdvEnt} & &\checkmark & 9.1 & 11.2 & 10.1 \\
     	    \cline{1-6}
                \cline{1-6}
     	CPN-low \cite{chen2018cascaded} & &\checkmark & 23.7 & 23.9 & 23.8 \\
     	CPN-well \cite{chen2018cascaded} & \checkmark& & 15.2 & 15.6 & 15.4 \\
     	CPN-all \cite{chen2018cascaded} &\checkmark &\checkmark & 32.8 & 31.7 & 32.2 \\
     	LLFlow \cite{LLFlow} &\checkmark &\checkmark & 25.6 & 28.2 & 27.0 \\
     	LIME \cite{LIME} &\checkmark & \checkmark& 33.2 & 28.4 & 30.7 \\
     	DANN \cite{ganin2016domain} &\checkmark &\checkmark & 27.9 & 30.6 & 29.3 \\
     	AdvEnt \cite{AdvEnt} &\checkmark &\checkmark & 28.2 & 29.0 & 28.6 \\                       LSBN+LUPI \cite{ExLPose_2023_CVPR} &\checkmark &\checkmark & {\ul 35.3} & {\ul 35.1} & {\ul 35.2}  \\ 
     	    \cline{1-6}
         
              \cline{1-6}
           Ours & &\checkmark & \textbf{39.1} & \textbf{36.2} & \textbf{37.6}  \\
     	   \cline{1-6}
    \end{tabularx}
    }
\label{tab:ocn}
\end{table*}

\paragraph{Evaluation Protocol} 
We follow the COCO protocol \cite{lin2014microsoft}, as used in LSBN+LUPI \cite{ExLPose_2023_CVPR}, to validate on ExLPose-test and ExLPose-OCN.
%
Average precision at various thresholds (i.e., $AP@0.5:0.95$) is reported for every subset.

\paragraph{Implementation Details}
DEKR-W32 \cite{DEKR2021} is adopted as our backbone network for the main teacher and the student. HigherHRNet \cite{cheng2020higherhrnet} is modified to be the complementary teacher. 
In the pre-training stage, the input image of our framework is cropped to $512\times512$. The data augmentations follow the DEKR paper setting, which include random rotation $[-30^\circ, 30^\circ]$, random scaling $[0.75, 1.5]$, random translation $[-40, 40]$ and random flipping. The training strategy of both teachers is the same with DEKR. The only difference is the loss weighting factors, the $\lambda_m = 0.03$ and $\lambda_c = 0.001$.

In the knowledge acquisition stage, the confidence score thresholds are set to be $s_m=0.9$ and $s_c = 0.5$ respectively for dual teachers. In training, we use Adam optimizer. The learning rate is set as $1e-4$. $\lambda_{\rm sup}$ and $\lambda_{unsup}$ are set to be 1.0. Our framework is trained on 4 RTX3090 GPUs (24GB of VRAM each) with a per-GPU batch size of 9 for the pre-training and 8 for the dual teacher knowledge acquisition stage. 
The training time for the dual-teacher setup is 139 seconds per epoch, which is comparable to the single-teacher setting at 103 seconds per epoch.
Additional information is provided in supplementary material. 

\paragraph{Baselines} To compare with existing image enhancement and domain adaptation methods, two SOTA image enhancement methods (RFormer, LLFlow) \cite{cai2023retinexformer, LLFlow} and two SOTA domain adaptation methods (UDA-HE, AdvEnt) \cite{kim2022unified,AdvEnt} are used as baselines. 
%
To ensure a fair comparison, we employ the same base human pose estimator (DEKR) \cite{DEKR2021} for all the aforementioned methods, as well as for our method, utilizing only well-lit ground-truth data during training.
Following the setup in ExLPose \cite{ExLPose_2023_CVPR}, we replace AdvEnt's semantic segmentation backbone with the same 2D pose estimator used in our method. 
UDA-HE is a model-agnostic domain-adaptive human pose estimation method.
We upgrade the backbone network (ResNet) in UDA-HE by HRNet, the same backbone used by our method and re-train AdaIN to ensure a fair comparison.
Note, UDA-HE is configured under a bottom-up paradigm, which is included as a bottom-up baseline in \cref{tab:exlpose_bottom_up}.
%
%
For a comprehensive comparison, we add the top-down baselines reported in LSBN+LUPI \cite{ExLPose_2023_CVPR}, which are trained with both well-lit and low-light ground-truth data (unpaired). CPN \cite{chen2018cascaded} is used as the base human pose estimator for these top-down baselines.  
Among the additional baselines \cite{ExLPose_2023_CVPR}, there is an additional domain adaptation method (DANN) \cite{ganin2016domain} and another image enhancement method (LIME) \cite{LIME}. 
Note that the training data of LSBN+LUPI \cite{ExLPose_2023_CVPR} is different from all the baselines because it relies on paired well-lit and low-light images with ground truths. 
All top-down baselines employ the same top-down method (CPN) \cite{chen2018cascaded} as the base human pose estimator. 

\begin{table}[t]
 \centering
 \begin{minipage}[t]{0.48\textwidth} 
  \centering
  \caption{Evaluation on ExLPose-test with a bottom-up baseline \cite{DEKR2021}, image enhancement \cite{cai2023retinexformer,LLFlow}, and domain adaptation \cite{kim2022unified,AdvEnt} methods. All the methods are fully-supervised by well-lit labels. The best is \textbf{bold}. The second best is {\ul underlined}. }
  \setlength{\tabcolsep}{5pt}
  \resizebox{\columnwidth}{!}{
    \begin{tabular}{l|cccc|c}
            \hline
            \hline
            \multirow{2}{*}{\textbf{Methods}}  & \multicolumn{5}{c}{\textbf{AP$^{\uparrow}$@0.5:0.95}} \\
            \cline{2-6}
              & LL-N & LL-H & LL-E & LL-A & WL \\
            \cline{1-6}
            Base-well \cite{DEKR2021}  & 1.3 & 0.0 & 0.0 & 0.8 &\uline{60.3} \\ 
            \cline{2-6}
          
            \cline{2-6}
             RFormer \cite{cai2023retinexformer} & {\ul 16.9} & 2.3 & 0.1 & 5.7 & 60.0 \\
            \cline{2-6}
              LLFlow \cite{LLFlow}  & 11.0 & 0.8 & {\ul 0.9} & 4.5 & 60.1 \\
             \cline{2-6}
             
             UDA-HE \cite{kim2022unified} & 16.4 & {\ul 3.6} & 0.1 & {\ul 7.4} & 53.7 \\
            \cline{2-6}
             AdvEnt \cite{AdvEnt}  & 4.6 & 0.7 & 0.0 & 2.0 & 56.4 \\
            \cline{1-6}
             \textbf{Ours} & \textbf{35.6} & \textbf{18.6} & \textbf{5.0} & \textbf{21.1} & \textbf{61.6} \\
            \cline{1-6}
    \end{tabular}
    }
  \label{tab:exlpose_bottom_up}
 \end{minipage}
 \hfill %
 \begin{minipage}[t]{0.48\textwidth} 
  \centering
  \caption{Evaluation on ExLPose-test with top-down methods trained on both low-light and well-lit labels (denoted by $\dag$). Differently, our method is a bottom-up method trained with well-lit labels alone. The best is \textbf{bold}. The second best is {\ul underlined}.}
  \setlength{\tabcolsep}{5pt}
  \resizebox{\columnwidth}{!}{
    \begin{tabular}{l|cccc|c}
            \hline
            \hline
             \multirow{2}{*}{\textbf{Methods}} &  \multicolumn{5}{c}{\textbf{AP$^{\uparrow}$@0.5:0.95}} \\
            \cline{2-6}
             & LL-N & LL-H & LL-E & LL-A & WL \\
            \cline{1-6}
            CPN-low$^\dag$ \cite{chen2018cascaded}  &  25.4 & 18.2 & 5.0 & 17.2 & 1.2 \\ 
            \cline{2-6}
             CPN-well$^\dag$ \cite{chen2018cascaded} &   15.3 & 2.7 & 0.4 & 6.7 & 59.9 \\
            \cline{2-6}
             CPN-all$^\dag$ \cite{chen2018cascaded} &   26.4 & 18.2 & {\ul 6.1} & 17.6 & 52.3 \\
            \cline{2-6}
            LLFlow$^\dag$ \cite{LLFlow} &  28.7 & 15.7 & 5.3 & 17.4 & {\ul 60.7} \\
            \cline{2-6}
            LIME$^\dag$ \cite{LIME} & {\ul 31.9} & \textbf{21.2} & \textbf{7.6} & \textbf{ 21.1} & 57.7 \\
             \cline{2-6}
            DANN$^\dag$ \cite{ganin2016domain} &  28.0 & 17.5 & 5.3 & {\ul 17.8} & 52.0 \\
            \cline{2-6}
            \cline{1-6}
            \textbf{Ours} &   \textbf{35.6} &  {\ul 18.6} &  5.0 & \textbf{21.1} & \textbf{61.6} \\
            \cline{1-6}
    \end{tabular}
    }
    \label{tab:exlpose_top_down}
 \end{minipage}
\end{table}

\subsection{Performance on ExLPose-OCN}
Quantitative comparison on ExLPose-OCN  is provided in \cref{tab:ocn}. 
The upper part of the table is the bottom-up baselines and the lower part is top-down. Note that our method is a bottom-up one. 
The top-down baselines include CPN fully supervised on low-light images (CPN-low), well-lit images (CPN-well), or both of them (CPN-all), CPN integrated with low-light image enhancement methods \cite{LLFlow,LIME}, and domain adaptation methods \cite{ganin2016domain,AdvEnt}. 
It is worth mentioning all the top-down baselines (except the CPN-well) are fully supervised by both the well-lit and low-light ground truths and the reported performance, all using ground-truth person detections. 
%
%
The bottom-up baselines include DEKR trained on labeled low-light or well-lit images, denoted as Base-low and Base-well, respectively. Also, we integrate SOTA image enhancement methods, including RFormer \cite{cai2023retinexformer} and LLFlow \cite{LLFlow}, as well as domain adaptation methods, including output-level method UDA-HE \cite{kim2022unified} and feature-level method AdvEnt \cite{AdvEnt}. 
The performance of the image enhancement methods is tested by applying the Base-well to the enhanced images. Therefore, all the bottom-up baselines including ours are trained with the well-lit ground-truth data alone with the training set of ExLPose and we only use ExLPose-OCN for evaluation.

Our method outperforms the SOTA method \cite{ExLPose_2023_CVPR} by 6.8\% (2.4 AP) on average. 
%
As ExLPose-OCN captures actual low-light images, the superior performance validates the effectiveness of our method in addressing the low-light challenges, where our method surpasses all the bottom-up and top-down baselines without using low-light ground truths or paired well-lit and low-light data. 
This highlights the potential of our method in real-world applications, where we achieve performance surpassing the SOTA methods without using low-light ground truths.

\subsection{Performance on ExLPose-test}
We compare our method with various bottom-up baselines on ExLPose-test, including a  fully-supervised baseline \cite{DEKR2021} and four baselines using well-lit ground-truth data only for training based on image enhancement \cite{LLFlow, cai2023retinexformer} or domain adaptation \cite{kim2022unified, AdvEnt}, as shown in \cref{tab:exlpose_bottom_up}. 
As expected, the fully-supervised baseline, Base-well, is heavily biased towards their training data due to the lack of cross-domain generalization. Its poor performance clearly demonstrates the substantial gap between well-lit and extremely low-light domains. 
Our method outperforms the baselines using image enhancement or domain adaptation. 
In particular, our performance beats the second best method on the LL-N subset by 18.7 AP which almost doubled the second-best. For LL-H and LL-E subsets, all methods except ours struggled to achieve reasonable performance. 
Our method also maintains the best performance on the WL (well-lit) subset compared with DEKR-well. 
%

To make a comprehensive comparison, we further compare with top-down baselines incorporated with image enhancement and domain adaptation methods in \cref{tab:exlpose_top_down}. 
Note that the comparison is not fair to our approach as the top-down baselines reported in \cite{ExLPose_2023_CVPR} were all trained using both low-light and well-lit ground truths. 
Furthermore, top-down human pose estimation methods generally outperform their bottom-up counterparts due to their utilization of human detection, as well as the reduction in human scale variation achieved through detection cropping and human patch normalization \cite{cheng2023bottom}.

However, we still achieve the highest performance on the LL-N, LL-A, and WL subsets in \cref{tab:exlpose_top_down}, while obtaining comparable performance on other subsets without utilizing the low-light ground-truth data. 
%
Note LSBN+LUPI \cite{ExLPose_2023_CVPR} is not included because its training data is different from the baselines in \cref{tab:exlpose_top_down}. 
LSBN+LUPI utilizes paired well-lit and low-light data and the ground truths, while the baselines included in the table use unpaired well-lit and low-light datasets in training. 
%
%
With the extra paired training data and ground truths, LSBN+LUPI manages to attain competitive performance on ExLPose-test, where its performance on the WL subset (61.5) is still worse than ours, but it is better than ours in LL-A (25.0).
However, when evaluating the generalization ability of each method on the real low-light dataset (ExLPose-OCN), where there is no training and only testing, our method surpasses all the baselines, including LSBN+LUPI, as demonstrated in \cref{tab:ocn}.

\begin{table}[t]
    \footnotesize
    \centering
    \caption{Ablation Study of the proposed modules in our method over ExLPose-test. PT stands for Pre-Training stage. Step 1 is the fully-supervised training on well-lit images. Step 2 is the fully supervised training with fake low-lit images after adding ELLA. KA indicates the Knowledge Acquisition stage. Single indicates the training with single teacher. Dual indicates our dual-teacher framework. The best is \textbf{bold}. The second best is {\ul underlined}. }
    \resizebox{0.57\columnwidth}{!}{
        \begin{tabular}{c|P{0.9cm}|P{0.9cm}|cccc|c}
            \hline
            \hline
            \multirow{2}{*}{} &   \multicolumn{2}{c|}{\textbf{Our Design}} & \multicolumn{5}{c}{\textbf{AP$^{\uparrow}$@0.5:0.95}} \\
             \cline{2-8}
             & ELLA & PDA & LL-N & LL-H & LL-E & LL-A & WL \\
             \cline{1-8}
            PT Step 1  &   &  & 1.3 & 0.0 & 0.0 & 0.8 & 60.3 \\
            \cline{2-8}
            PT Step 2  &   \checkmark &  & 29.4 & 13.6 & 1.6 & 15.8 & \textbf{62.1} \\
            \cline{2-8}
            KA &   \checkmark &  & 29.5 & 14.6 & 2.9 & 16.8 & 60.5 \\
            \cline{2-8}
            Single  &  \checkmark & \checkmark & {\ul 33.4} & {\ul 16.6} & {\ul 3.1} & {\ul 19.0} & 59.4 \\
            \cline{2-8}
            Dual &   \checkmark & \checkmark & \textbf{35.6} & \textbf{18.6} & \textbf{5.0} & \textbf{21.1} & {\ul 61.6} \\
            \cline{1-8}
        \end{tabular}
    }
    \label{tab:ablation} 
\end{table}

\begin{table}[t]
    \footnotesize
    \centering
    \caption{Number of the additional pseudo labels provided by the complementary teacher in ExLPose training set. OKS Thres., referring to OKS threshold, is used to determine the validity of pseudo labels by comparing the OKS between pseudo labels and the ground truth with them. PL indicates pseudo labels. Main indicates the mean teacher. Comp. indicates complementary teacher.}
    \resizebox{0.75\columnwidth}{!}{
        \begin{tabular}{c|c|c|c}
            \hline
            \hline
            OKS Thres. & \# of PL in Main & \# of PL in Comp. & \# of additional PL \\
            \cline{1-4}
               0.3   &  10203  &     12155    &     4870 (+47.7\%)\\
               0.5   &  5056   &      8735    &     2136 (+49.8\%)\\
               0.7   &  3293   &      5435    &     1180 (+35.8\%)\\
               0.9   &  1463   &      1574    &     501  (+34.2\%)\\

            \cline{1-4}
        \end{tabular}
    }
    \label{tab:dual_teach_pseudo_labels_statis}
\end{table}

\subsection{Ablation Studies}
Ablation study is performed to validate the effectiveness of the individual modules in our method is shown in \cref{tab:ablation}. 
We use our framework in the pre-training stage without any newly proposed components as the baseline, and subsequently add each new module in separate experiments.

\paragraph{ELLA} We evaluate our method's performance with and without the supervision of fake low-light images generated from WL. The results are in the first two rows of \cref{tab:ablation}. The inclusion of ELLA significantly boosts the performance on LL-N, LL-H, and LL-A test subsets from almost zero to a reasonable level. 

\paragraph{PDA} 
We investigate our method's performance with and without PDA in Stage 2 training. 
When comparing the third row (single teacher without PDA) to the fourth row (single teacher with PDA), we observe a $20.2\mathcal{\%}$ (3.4 AP) improvement on LL-A, accompanied by a marginal 0.4 AP drop on WL. This highlights the effectiveness of PDA in enabling the student to surpass the teachers.

\paragraph{Dual Teacher}
Comparing the last two rows, we observe that the model's performance with dual teachers (last row) surpasses that of a single teacher (second last row) in every split. 
Notably, 
The complementary teacher achieves a $11\mathcal{\%}$ (2.1 AP) improvement, which is comparable to the $13\mathcal{\%}$ (2.2 AP) improvement obtained by the main teacher on LL-A (all subsets), underscoring the effectiveness of our dual-teacher framework. Additionally, with the help of the dual teachers, the student model exhibits enhanced performance on the well-lit split.

To validate that the complementary teacher provides more valid pseudo labels, quantitative evaluation is  provided in Table \ref{tab:dual_teach_pseudo_labels_statis}. We measure the validity of pseudo labels by computing their OKS with corresponding ground truths and count the number of non-overlapping pseudo labels by thresholding the OKS at 0.5 between the dual teachers, following \cite{lin2014microsoft, li2019crowdpose}. The evaluation demonstrates that the complementary teacher consistently produces additional valid pseudo labels across different thresholds of quality, providing an average of 41.8\% more valid pseudo labels compared to using the main teacher alone. The increased number of valid pseudo labels leads to further improvement in the student's training, highlighting the effectiveness of our dual-teacher framework.

\section{Conclusion}
We introduce the first human pose estimation method for extremely low-light conditions, utilizing well-lit ground-truth data exclusively.
Our novelty lies in leveraging center-based and keypoint-based teachers to generate enriched pseudo labels, enabling the student model to achieve competitive performance on extremely low-light images without the need for training with low-light ground truths. To our knowledge, this represents a significant advancement in the field of low-light human pose estimation.


\section*{Acknowledgements}
This research is in part supported by the Singapore Ministry of Education Academic Research Fund Tier 1 (WBS: A-8001229-00-00), a project titled ``Towards Robust Single Domain Generalization in Deep Learning''.

%
%
\bibliographystyle{splncs04}
\bibliography{main}
\end{document}


\title{Domain-Adaptive 2D Human Pose Estimation via Dual Teachers in Extremely Low-Light Conditions \\ -Supplementary Material-} 

\titlerunning{Abbreviated paper title}

\author{Yihao Ai\inst{1}\thanks{~Equal contribution.}\orcidlink{0009-0005-2336-4813}\and
Yifei Qi \inst{1}$^\star$\orcidlink{0009-0007-4197-947X}\and
Bo Wang\inst{2,3}\orcidlink{0000-0002-0127-2281}\and
Yu Cheng\inst{1} \and
Xinchao Wang\inst{1} \and
Robby T. Tan\inst{1}}

\authorrunning{Y.~Ai et al.}

\institute{Department of Electrical and Computer Engineering, National University of Singapore \and 
Department of Computer and Information Science, University of Mississippi \and CtrsVision \\ \email{\{yihao, e0315880\}@u.nus.edu}, hawk.rsrch@gmail.com, e0321276@u.nus.edu, \{xinchao, robby.tan\}@nus.edu.sg}

\maketitle


In this supplementary material readers can find: ($\rm \Rmnum{1}$) Additional ablation study on the dual teacher framework in \cref{sec:dual_ablation}. ($\rm \Rmnum{2}$) Discussion on the failure cases and limitations in \cref{sec:failure_cases_limitations}. \edit{}{($\rm \Rmnum{3}$) Discussion on the effectiveness of the style transfer. ($\rm \Rmnum{4}$) An additional case study in low-light conditions.} ($\rm \Rmnum{5}$) Additional qualitative examples on ExLPose-OCN, ExLPose-test\cite{ExLPose_2023_CVPR} in \cref{sec:addit_qual}. ($\rm \Rmnum{6}$) Additional qualitative examples on real nighttime dataset, NightOwl \cite{neumann2019nightowls} in \cref{sec:addit_nightowl}. ($\rm \Rmnum{7}$) Additional implementation details in \cref{sec:addit_imple}.

\section{Additional Dual-Teacher Ablation Studies}
\label{sec:dual_ablation}
The student model follows the main teacher who represents a human pose by a center and offsets between the center and each keypoint. 
Specifically, the center is defined as the average of the keypoints, and the offsets are defined as a set of 2D vectors that point from the center to each keypoint.
In the second stage (dual-teacher knowledge acquisition stage) of our method, both teachers are supposed to teach the student with their pseudo labels. 
%
The main teacher directly predicts the center for each individual, but the complementary teacher predicts each keypoint directly, and the average of the keypoints of a person is used as the human center for the individual. 
%
In this situation, the accuracy of the human centers from the pseudo labels from both teachers plays an important role in the student's learning, because the accuracy of the human centers affects the keypoints of an individual as the offsets to each keypoint are predicted based on a given human center. 
Therefore, to understand and verify the role that each teacher plays in the second stage training, it is important to assess the quality of the human center predictions from both teachers.
\begin{minipage}[c]{\textwidth}
  \begin{minipage}[l]{0.49\textwidth}
    \includegraphics[width=1.0\linewidth]{figures/center_dis.pdf}
    \vspace{-0.8cm}
    \captionof{figure}{The error distributions of the main and complementary teacher's center prediction.}
    \label{fig:center_error_dis}
  \end{minipage}
  \hfill
  \begin{minipage}[c]{0.49\textwidth}
    \centering
    \vspace{-0.8cm}
    \captionof{table}{Center statistics of the main and complementary teachers. Main stands for the main teacher. Comp. denotes the complementary teacher. Average error indicates the mean of the error distribution for each teacher. Variance is calculated individually from the error distribution of each teacher.}
    \resizebox{0.7\columnwidth}{!}{
        \begin{tabular}{c|c|c}
            \hline
            \hline
                &  average error & variance \\
            \cline{1-3}
            Main & 3.43 & 6.26\\
            Comp. & 7.00 & 11.99\\
            \cline{1-3}
        \end{tabular}
    }
    \label{tab:center_statistics} 
    \end{minipage}
\end{minipage}

To this end, we provide the analysis of the predicted human centers from both teachers in {\cref{tab:center_statistics}} and {\cref{fig:center_error_dis}}. 
In particular, we compare the ground-truth human centers and the predicted ones of both teachers. 
If the pixel distance between a ground-truth human center and a predicted center from a teacher is less than 20 pixels, we count the person with this ground-truth center as captured (or matched), and the pseudo label (human pose) based on the predicted human center can be used.
%
%
A pixel-level error between the ground-truth center and the matched teacher's center is calculated in \cref{tab:center_statistics}, aiming to compare which teacher produces accurate centers.
The average error and the variance of errors are provided in the first and second columns in \cref{tab:center_statistics}. The corresponding error distributions of the two teachers are shown in \cref{fig:center_error_dis}. We observe the main teacher produces more accurate human center prediction than the complementary teacher, with smaller average error and variance. 
Combining the evaluation shown in Tab. 5 of the main paper that the complementary teacher produces more additional pseudo labels compared to using the primary teacher alone, we demonstrate that the two teachers are complementary to each other, where the main teacher provides a small number of accurate pseudo labels, and the complementary teacher provides more pseudo labels but with lower quality. 

%

In \cref{fig:center_error_dis}, the major errors of the main teachers are concentrated at 3.5 (or near 0.0), while the errors of the complementary teacher are almost equally distributed between 3.5 and 8.5 and contain a long tail between 8.5 to 20.0.
This further illustrates the center prediction of the complementary teacher is relatively inaccurate and unstable, which may affect the training of the student.
This triggers us to look into using the complementary teacher alone in the training of the student model, which is provided in \cref{tab:ablation_comp}.  

\begin{table}[t]
    \footnotesize
    \centering
     \caption{Additional ablation studies of the proposed modules in our method over ExLPose-test in the knowledge acquisition stage. Single (main) indicates the training with the main teacher only. Single (comp.) indicates the training with the complementary teacher only. Dual indicates our dual-teacher framework. The best is \textbf{bold}. The second best is {\ul underlined}.}
    \resizebox{0.7\columnwidth}{!}{
        \begin{tabular}{c|P{0.9cm}|P{0.9cm}|c|c|c|c|c}
            \hline
            \hline
            \multirow{2}{*}{} &   \multicolumn{2}{c|}{\textbf{Our Design}} & \multicolumn{5}{c}{\textbf{AP@0.5:0.95}} \\
             \cline{2-8}
             & ELLA & PDA & LL-N & LL-H & LL-E & LL-A & WL \\
             \cline{1-8}
            Single (main) &  \checkmark & \checkmark & {\ul 33.4} & {\ul 16.6} & {\ul 3.1} & {\ul 19.0} & 59.4 \\
            Single (comp.) &  \checkmark & \checkmark &  32.0 & 15.4 & 2.5 & 18.0 &  {\ul 61.5}\\
            \cline{1-8}
            Dual (our full model) &  \checkmark & \checkmark & \textbf{35.6} & \textbf{18.6} & \textbf{5.0} & \textbf{21.1} &  \textbf{61.6} \\
            \cline{1-8}
        \end{tabular}
    }
    
    \label{tab:ablation_comp} 
\end{table}

We conduct an additional ablation study of using the complementary teacher alone as shown in the second row of \cref{tab:ablation_comp}.
It is observed that the student still improves after being taught by the complementary teacher solely, the performance is not as good as being taught by the main teacher alone as shown in the first two rows of \cref{tab:ablation_comp}. 
This additional ablation study demonstrates that the complementary teacher helps to enhance the student's performance, where the extent of improvement is limited by the quality of pseudo labels generated by the complementary teacher alone.
Using either the main or the complementary teacher alone does not yield optimal results.
By taking advantage of the pseudo labels from both teachers, the best performance is achieved as shown in the last row of \cref{tab:ablation_comp}. 

It is noteworthy that our performance on LL-A (21.1 AP), as presented in last row of \cref{tab:ablation_comp}, is comparable to that of Base-low (22.4 AP), which was trained exclusively with low-light ground truths. 
As a supervised method, Base-low experiences significant overfitting to low-light images, achieving nearly zero performance (0.3 AP) on the well-lit split (WL). 
In contrast, our method maintains robust performance in well-lit conditions (61.6 AP), underscoring our approach's effectiveness. 
Notably, Base-low was excluded from Table 2 in the main paper to ensure a fair comparison among methods not utilizing low-light ground truths.

\section{Failure Cases and Limitations}
\label{sec:failure_cases_limitations}
\begin{figure}[ht]
	\centering
    \begin{minipage}[c]{0.159\linewidth}
		\centerline{\scriptsize{Input Image}}
		\vspace{0.3em}
	\end{minipage}
    \begin{minipage}[c]{0.159\linewidth}
		\centerline{\scriptsize{Ours}}
		\vspace{0.3em}
	\end{minipage}
	\begin{minipage}[c]{0.159\linewidth}
		\centerline{\scriptsize{Ground Truth}}
		\vspace{0.3em}
	\end{minipage}
    \begin{minipage}[c]{0.159\linewidth}
		\centerline{\scriptsize{Input Image}}
		\vspace{0.3em}
	\end{minipage}
    \begin{minipage}[c]{0.159\linewidth}
		\centerline{\scriptsize{Ours}}
		\vspace{0.3em}
	\end{minipage}
	\begin{minipage}[c]{0.159\linewidth}
		\centerline{\scriptsize{Ground Truth}}
		\vspace{0.3em}
	\end{minipage}

    \begin{minipage}[c]{0.159\linewidth}
		\includegraphics[width=\linewidth]{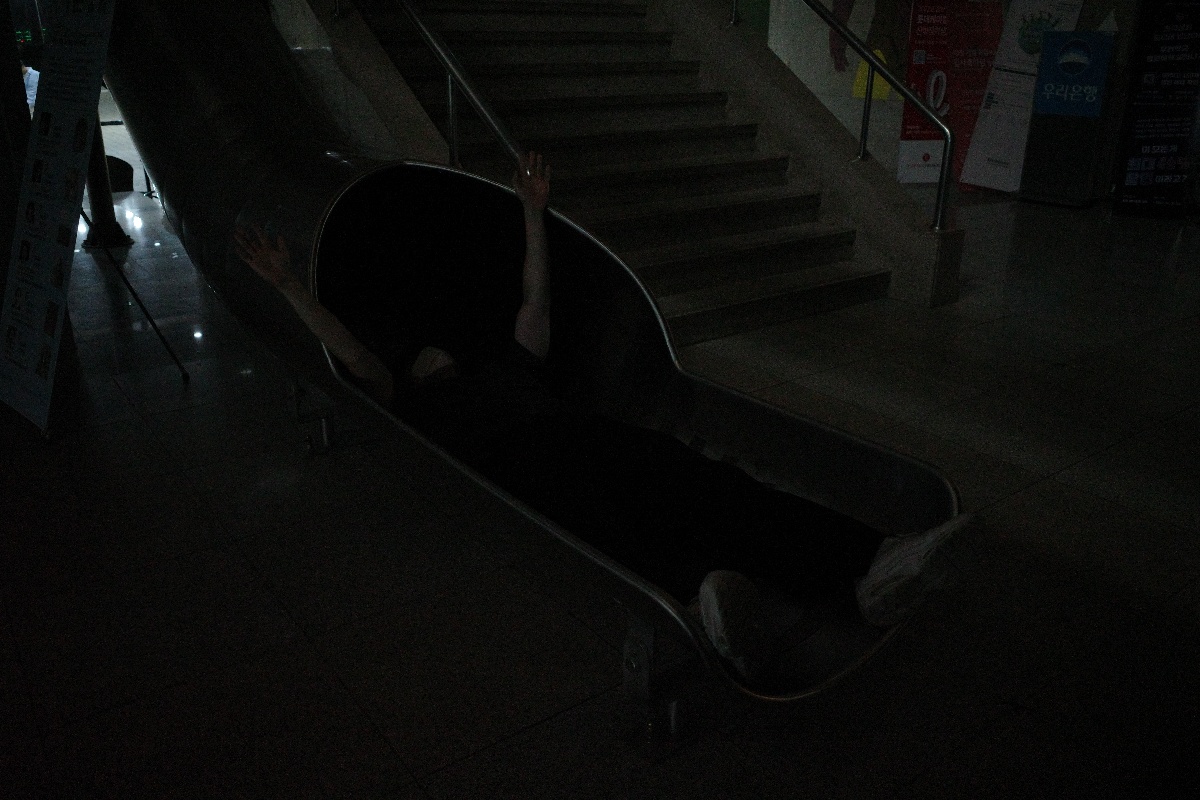}
	\end{minipage}
	\begin{minipage}[c]{0.159\linewidth}
		\includegraphics[width=\linewidth]{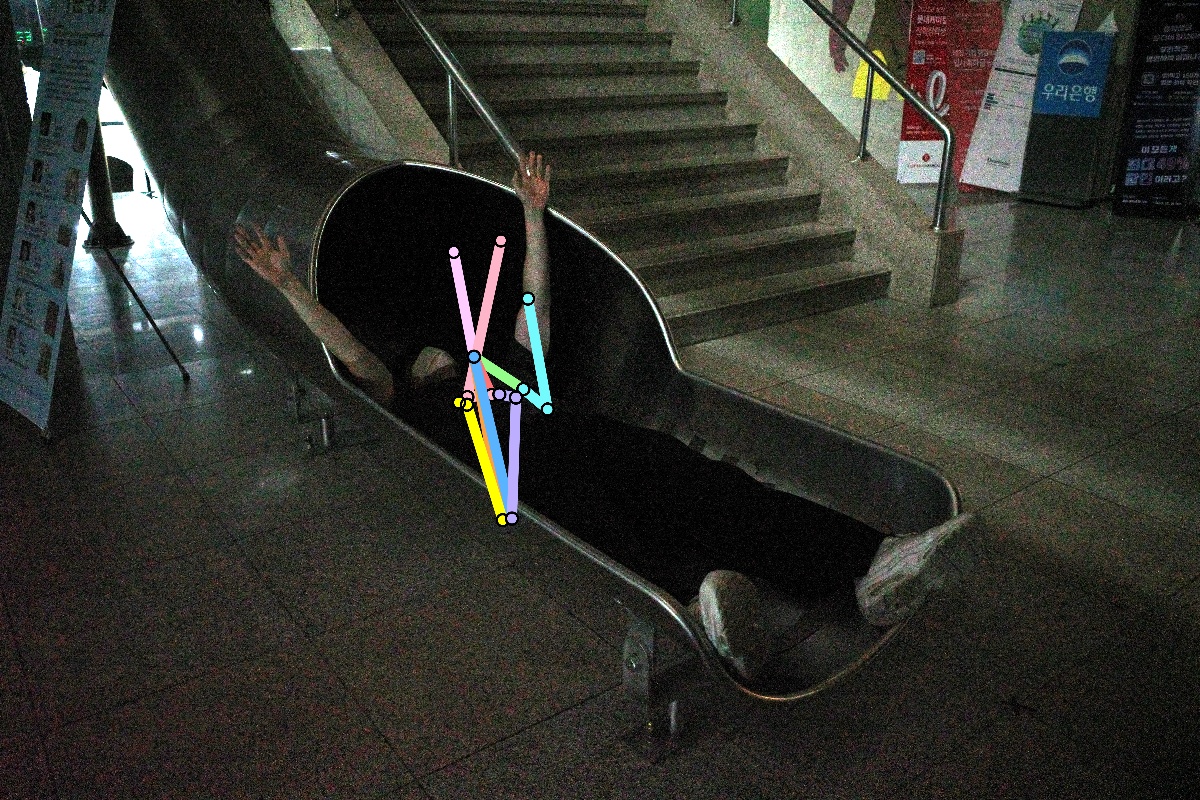}
	\end{minipage}
       \begin{minipage}[c]{0.159\linewidth}
		\includegraphics[width=\linewidth]{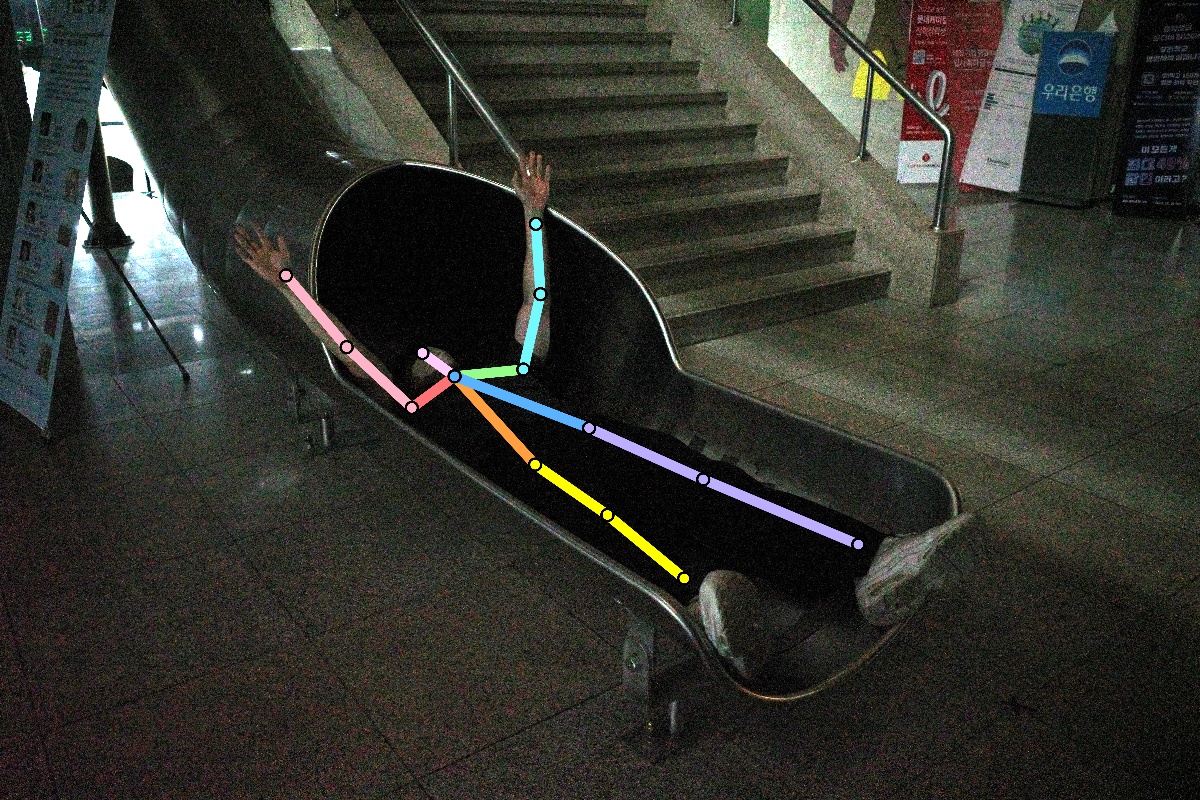}
	\end{minipage}
    \begin{minipage}[c]{0.159\linewidth}
		\includegraphics[width=\linewidth]{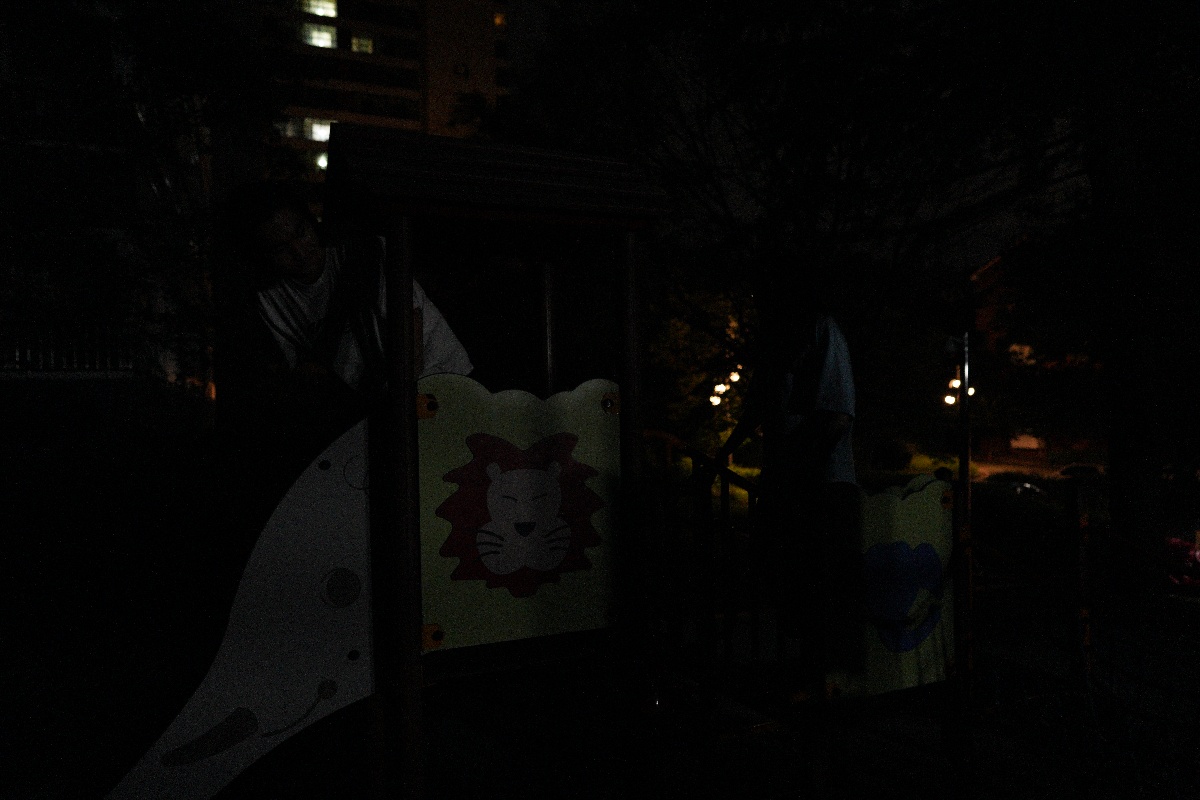}
	\end{minipage}
    \begin{minipage}[c]{0.159\linewidth}
		\includegraphics[width=\linewidth]{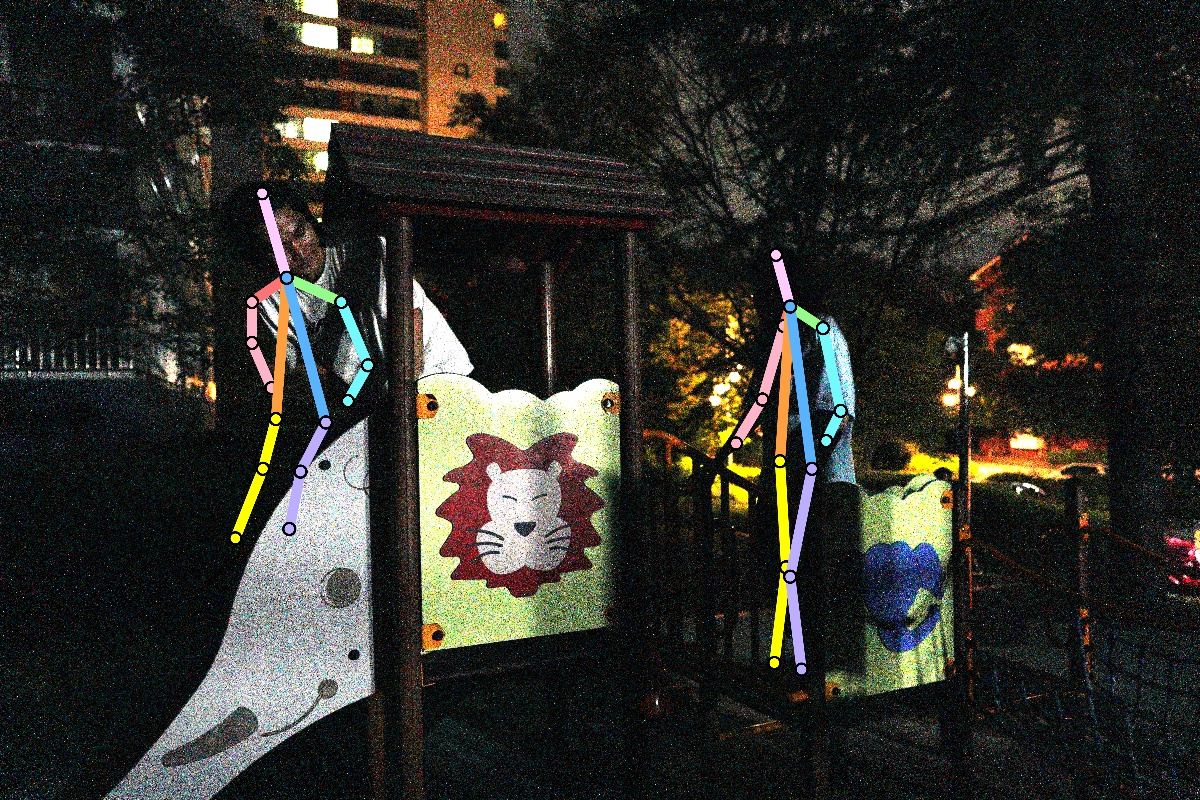}
	\end{minipage}
       \begin{minipage}[c]{0.159\linewidth}
		\includegraphics[width=\linewidth]{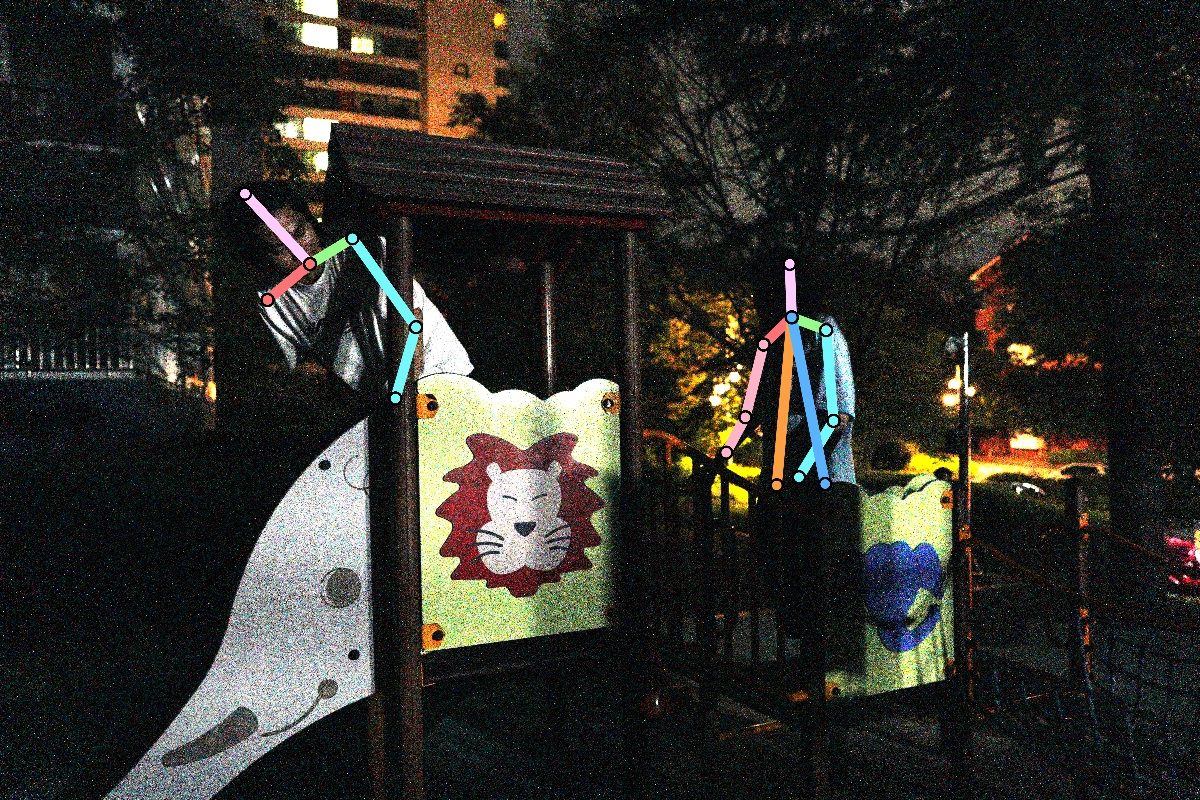}
	\end{minipage}
    
	\caption{Failure cases of the proposed method. 
    The first and fourth images show the original extremely low-light images, which serve as input for our method. The other images have been brightened solely for visualization purposes. The brightened images are not utilized by our method. 
    }
	\label{fig:failure cases}
	\vspace{-0.2cm}
\end{figure}

Our method may produce inaccurate predictions for rare human poses and object occlusion. 
As depicted in the first three images of \cref{fig:failure cases}, our method fails to accurately estimate the pose when a person is lying down. This discrepancy arises because the dataset is primarily dominated by standing individuals. 
In the last three images, when the left person's body is partially occluded by a board, our method incorrectly predicts the keypoints for the lower body. 
Note that these invisible keypoints are not annotated in the ground truth.

The ExLPose dataset \cite{ExLPose_2023_CVPR} may present certain limitations. 
The primary concern is that its well-lit split is relatively smaller than commonly used daytime datasets \cite{lin2014microsoft, li2019crowdpose}. 
With a larger set of well-lit images containing a strong prior on human pose, the challenge of low-light human pose estimation may not be as distinct, potentially reducing the requirement of specific techniques to handle this challenge. 
However, additional experiments indicate that incorporating the larger set of well-lit images \cite{li2019crowdpose} has minimal impact on low-light conditions. 
The experimental results are provided in \cref{tab:r3_exp}, where we use CrowdPose to replace the ExLPose well-lit split in training.
In \cref{tab:r3_exp}, the model trained on CrowdPose demonstrates improved performance (68.9 AP) on the WL split in the last column, compared with the model trained on ExLPose well-lit split (61.6 AP). 
This is reasonable as CrowdPose contains more diverse well-lit data than that of the ExLPose well-lit split. 
However, a detailed examination of all the low-light splits (i.e., LL-N, LL-H, LL-E, and LL-A) indicates that the model trained on CrowdPose does not exhibit enhanced performance in low-light conditions.
Therefore, merely employing a larger or more diverse dataset of well-lit images fails to address the challenges of low-light human pose estimation, underscoring the need for a dedicated approach tailored to low-light scenarios.  
%

\begin{table}[t]
    \footnotesize
    \centering
    \caption{Evaluation on ExLPose-test with DEKR \cite{DEKR2021}. PT and KA denote pre-training stage and knowledge acquisition stage. $\dag$ indicates the model is trained with CrowdPose \cite{li2019crowdpose}. The best is \textbf{bold}. The second best is {\ul underlined}. }
    \begin{tabular}{c|c|c|c|c|c}
            \hline
            \hline
            \textbf{Methods} & LL-N & LL-H & LL-E & LL-A & WL \\
            \cline{1-6}
            PT$^\dag$ & 30.0 & 13.3  & 3.4 & 16.5 &  {\ul 64.8}\\
            KA$^\dag$ & {\ul 35.3} & {\ul 16.6} & {\ul 3.5} & {\ul 19.6} & \textbf{68.9} \\
            \cline{1-6}
            Ours & \textbf{35.6} & \textbf{18.6} & \textbf{5.0} & \textbf{21.1} & 61.6 \\
            \cline{1-6}
    \end{tabular}
    \label{tab:r3_exp}
    \vspace{-0.5cm}
\end{table}

\section{Effectiveness of Style Transfer} 
\edit{}{The inferior performance of UDA-HE \cite{kim2022unified} is due to the inadequate quality of style transfer. Specifically, we present qualitative examples from the style transfer network AdaIN \cite{huang2017arbitrary}, utilized in UDA-HE \cite{kim2022unified}. The images generated from well-lit scenes, derived from low-light conditions, demonstrate significant detail loss or distortion, as illustrated in Figures (a) to (c) of \cref{fig:style_transfer_sample}.} 
\edit{}{Conversely, unnatural artifacts are evident in human subjects of low-light images transformed from well-lit input, shown in (d) to (f) of \cref{fig:style_transfer_sample}.}

\begin{figure}[!htb]
    \vspace{-0.2cm}
    \begin{subfigure}{0.32\linewidth}
        \includegraphics[width=\linewidth]{figures/style_transfer_samples/LL-WL/Content_LL.png}
        \caption{Low-light input}
    \end{subfigure}
    \begin{subfigure}{0.32\linewidth}
        \includegraphics[width=\linewidth]{figures/style_transfer_samples/LL-WL/Reconstructed_WL.png}
        \caption{Well-lit output}
    \end{subfigure}
    \begin{subfigure}{0.32\linewidth}
        \includegraphics[width=\linewidth]{figures/style_transfer_samples/LL-WL/Style_WL.png}
        \caption{Well-lit GT}
    \end{subfigure}

    \begin{subfigure}{0.32\linewidth}
        \includegraphics[width=\linewidth]{figures/style_transfer_samples/WL-LL/Content_WL.png}
        \caption{Well-lit input}
    \end{subfigure}
    \begin{subfigure}{0.32\linewidth}
        \includegraphics[width=\linewidth]{figures/style_transfer_samples/WL-LL/Reconstructed_LL.png}
        \caption{Low-light output}
    \end{subfigure}
    \begin{subfigure}{0.32\linewidth}
        \includegraphics[width=\linewidth]{figures/style_transfer_samples/WL-LL/Style_LL.png}
        \caption{Low-light GT}
    \end{subfigure}
    \caption{Samples of the style transfer net AdaIN \cite{huang2017arbitrary}. The first row illustrates the process of transferring a low-light image to a well-lit image. The second row illustrates the process of transferring a well-lit image to a low-light image.}
    \label{fig:style_transfer_sample}
\end{figure}

\section{Case study in low-light conditions}

\edit{}{\cref{tab:factors} presents a case study of first stage training in low-light conditions on ExLPose-test.
"Base" indicates the performance of applying brightness adjustment alone to well-lit images to ensure basic low-light conditions.
Subsequent rows show the performance of adding motion blur, low image contrast or high-level noise respectively.
In particular, motion blur is simulated following 2PCNet [25] by adding Gaussion blur with direction. 
%
Adding noise brings the largest improvement, as it impacts the pixel-level accuracy of keypoint localization. Low image contrast also helps but it is less effective than noise.
Motion blur simulation yields minor improvement, as ExLPose [29] contains few individuals with motion blurring issues. However, it could potentially play an important role in low-light video data, where motion blur is a major source of noise. 
}
\begin{table}[t]
    \footnotesize
    \centering
    \caption{Study of different factors (blur, contrast, and noise) in the low-light conditions.}
    \begin{tabular}{l|c|c|c|c|c}
            \hline
            \hline
            \textbf{Methods} & LL-N & LL-H & LL-E & LL-A & Improv. \\
            \cline{1-6}
            Base & 15.9 & 1.9 & 0.1 & 6.6 & N/A \\
            +Blur & 16.5 & 3.0 & \underline{0.2} & 7.1 & 0.5 \\
            +Contrast & \underline{20.1} & \underline{5.0} & \underline{0.2} & \underline{9.0} & \underline{2.4} \\
            +Noise & \textbf{25.0} & \textbf{11.5} & \textbf{1.7} & \textbf{13.4} & \textbf{6.8} \\
            \cline{1-6}
    \end{tabular}
    \label{tab:factors}
    \vspace{-0.5cm}
\end{table}

\section{Additional Qualitative Results}
\label{sec:addit_qual}
\subsection{ExLPose-OCN}

Additional qualitative results on ExLPose-OCN \cite{ExLPose_2023_CVPR} are presented from {\cref{fig:exlocn_supp_1}} to \cref{fig:exlocn_supp_5}, where it is observed that our method significantly outperforms the baseline methods \cite{AdvEnt,LLFlow,kim2022unified,cai2023retinexformer} on the real extremely low-light images from ExLPose-OCN. 
When individuals are dark or indistinguishable from their background as shown in the 1st sample of {\cref{fig:exlocn_supp_1}} and the 2nd sample of {\cref{fig:exlocn_supp_4}}, our method is the only one that produces reasonable human pose estimation in such challenging cases. 
In the multi-person situation, such as the 2nd sample of {\cref{fig:exlocn_supp_2}} and the 2nd sample of \cref{fig:exlocn_supp_5}, only our method provides the human pose estimation for each person. 
For the relatively visible samples, such as the 1st sample of {\cref{fig:exlocn_supp_3}}, our method provides more accurate human pose estimation than the baseline methods (e.g., the two individuals in front).

\subsection{ExLPose-test}

We provide qualitative results on ExLPose-test \cite{ExLPose_2023_CVPR} from {\cref{fig:exltest_supp_1}} to {\cref{fig:exltest_supp_5}} to demonstrate the superior performance of our method on the low-light images from ExLPose-test compared with the baseline methods \cite{AdvEnt,LLFlow,kim2022unified,cai2023retinexformer}. 
In general, the baseline methods demonstrate poor performance on the low-light images from ExLPose-test. In contrast, our method consistently produces accurate low-light human pose estimations for these images.
The baseline methods exhibit severe limitations on extremely low-light samples, such as those in the 2nd sample of {\cref{fig:exltest_supp_2}} and 1st sample of \cref{fig:exltest_supp_5}. This performance aligns with their overall poor results on the LL-H and LL-E subsets, as demonstrated in Tab. 2 of the main paper.
For individuals who are relatively more visible (e.g., the 1st sample of {\cref{fig:exltest_supp_3}} and the 2nd sample of \cref{fig:exltest_supp_5}), the baseline methods produce some human pose predictions, but they are less accurate or wrong. In contrast, our method outperforms in generating more accurate human pose estimations.

\section{Additional Results on Nighttime Dataset}
\label{sec:addit_nightowl}
To validate our method on other datasets with poor lighting conditions other than ExLPose-OCN and ExLPose-test, we provide the additional qualitative results on a nighttime dataset, NightOwl \cite{neumann2019nightowls}. 
The nighttime condition is different from but closely related to the low-light scenarios.
Both types of images are taken in the dark. Yet the nighttime images contain man-made light sources, such as lamps, LEDs, and car lights, while the low-light images are free of light sources at all.
%
%
%
%
The NightOwl dataset focuses on pedestrian detection at night, therefore containing various human objects in real nighttime conditions.
However, only bounding boxes of objects are annotated as this dataset is proposed for object detection in the nighttime.
Without human pose annotations, we provide the qualitative results to validate our method compared with the baseline methods \cite{AdvEnt,LLFlow,kim2022unified,cai2023retinexformer} as shown in \cref{fig:nightowl_1}. 
Note, the nighttime dataset is only used for testing purposes, no training or fine-tuning has been performed on any of the methods in \cref{fig:nightowl_1}.

In general, our method produces pose estimations for visible human objects and relatively dark human objects simultaneously.
For example, the middle person in the 1st sample of \cref{fig:nightowl_1}, our method and UDA-HE \cite{kim2022unified} provide a predicted human pose.  
However, for the low-visibility persons in the left part of the image, only our method successfully predicts the human poses.
In the 2nd sample of \cref{fig:nightowl_1}, the baseline methods struggle with predicting human poses for low-contrast individuals, while our method effectively provides pose estimations in low-light conditions.
The additional results on the nighttime dataset demonstrate our method's ability to handle challenging lighting conditions, including nighttime scenarios.
This further validates the effectiveness of our method in human pose estimation under poor lighting conditions, extending its applicability beyond low-light scenarios.

\section{Additional Implementation Details}
\label{sec:addit_imple}
\paragraph{Loss Function} In the main teacher, we use MSE loss and smooth L1 loss for optimizing the heatmaps (including the $K$ keypoint heatmaps and one center heatmap) and offsets, which are denoted as $L_H^{m}$ and $L_O^{m}$ respectively in the main paper. Two losses are defined by the following \cref{equ:main_heatmap_loss} and \cref{equ:main_offset_loss}:
\begin{equation}
\scalebox{0.9}{
	$L_H^{m} = \frac{1}{K+1}\sum_{i=1}^{K+1}||W_i(H-\hat{H})||_2^2,$
}
\label{equ:main_heatmap_loss}
\end{equation}
\begin{equation}
\scalebox{0.9}{
	$L_O^{m} = \sum_{i\in{\psi}}\frac{1}{\sqrt{h_i^2+w_i^2}}\mathrm{Smooth_{L1}}(O_i-\hat{O}_i),$
}
\label{equ:main_offset_loss}
\end{equation}
where $H \in{\mathbb{R}^{(K+1)\times{h}\times{w}}}$ and $O \in{\mathbb{R}^{(2K)\times{h}\times{w}}}$ stands for the ground-truth heatmaps and offset maps. $\hat{H}$ and $\hat{O}$ represent the predicted heatmaps and offset maps of the main teacher. $W$ is the weight map for the heatmaps, $\psi$ is the set of keypoints with labels, and $h_i$ and $w_i$ are the height and width of the $i$-th person's bounding box. 

In the complementary teacher, the heatmap loss $L_H^{c}$ is the same as $L_H^{m}$ except that there is no center heatmap. The tag map loss $L_{\rm tag}^{c}$ is defined by the following equation:
\begin{equation}
\scalebox{0.88}{
    $L_{\rm tag}^{c} = \frac{1}{N}\sum_{i}\sum_{k}(\bar{t}_i - t_k^i)^2 +       
    \frac{1}{N^2}\sum_{i}\sum_{i^\prime}\exp\{{-\frac{1}{2\sigma^2}(\bar{t}_i - \bar{t}_{i^\prime})^2}\}.$
}
\end{equation}
where $t_k^i$ stands for the tag of the $k$-th keypoints of the $i$-th person that retrieved in the tag map. $\bar{t}_i$ represents the average tag of the $i$-th person, i.e. $\frac{1}{K}\sum_{k}t_k^i$. $N$ denotes the total number of persons. $\sigma$ is set to be $\frac{\sqrt{2}}{2}$. 
The supervised loss of the complementary teacher $L_{\rm sup}^{c}$ is a linear combination of the heatmap loss and the tag map loss (similar to Eq. (2) in the main paper), which is defined by the following \cref{equ:lsup_comp}:
\begin{equation}
\scalebox{0.9}{
	$L_{\rm sup}^{c} = L_H^{c} + \lambda_c L_{\rm tag}^{c}$
}
\label{equ:lsup_comp}
\end{equation}
%

\paragraph{Additional Training Details}  In the pre-training stage, we train both teachers for 300 epochs on the well-lit images from ExLPose \cite{ExLPose_2023_CVPR}, and their augmentations (fake low-light images with ELLA). 
The learning rate is set to 1e-3. We use a learning rate scheduler to decay the learning rate by a factor of 0.1 at the $90$-th and $120$-th epochs. 
In the knowledge acquisition stage, the proposed PDA darkens the cropped image patch following Eq. (4) in the main paper (i.e., $I_{\rm out} = bI_{\rm in}$). 
In particular, $b$ is randomly selected from the range between 0.1 and 0.5, leading to less darkening on the real low-light images compared to the well-lit images darkened by augmentations in ELLA, considering that the real low-light image is already dark. 
%
%
Apart from the low-light specific augmentation, input images are also augmented by random affine transformations similar to DEKR \cite{DEKR2021}. 
The training is set to be 10 epochs with fixed teachers.

\paragraph{Code}
We have included the code of our method as part of the supplementary material in the .zip file. We will make our code and pre-trained model publicly available upon publication. 

\newpage

\input{figures/supp_fig_tex/exlocn/1}
\input{figures/supp_fig_tex/exlocn/2}
\input{figures/supp_fig_tex/exlocn/3}
\input{figures/supp_fig_tex/exlocn/4}
\input{figures/supp_fig_tex/exlocn/5}

\input{figures/supp_fig_tex/exltest/1}
\input{figures/supp_fig_tex/exltest/2}
\input{figures/supp_fig_tex/exltest/3}
\input{figures/supp_fig_tex/exltest/4}
\input{figures/supp_fig_tex/exltest/5}

\input{figures/nightowl_samples_supp_1}

%
%
\bibliographystyle{splncs04}
\bibliography{main}

%% file: figures/rep_samples.tex
\begin{center}
    \centering
    \begin{minipage}[c]{0.159\linewidth}
		\includegraphics[width=\linewidth]{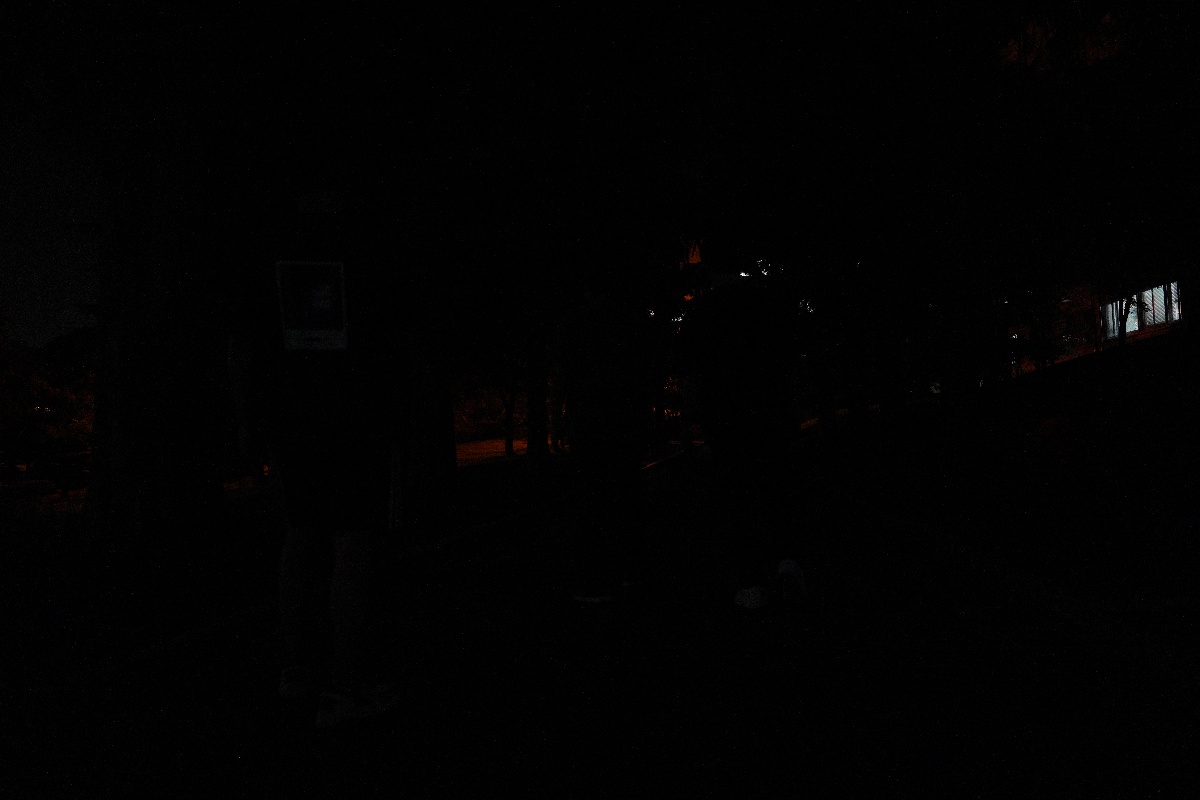}
	\end{minipage}
    \begin{minipage}[c]{0.159\linewidth}
		\includegraphics[width=\linewidth]{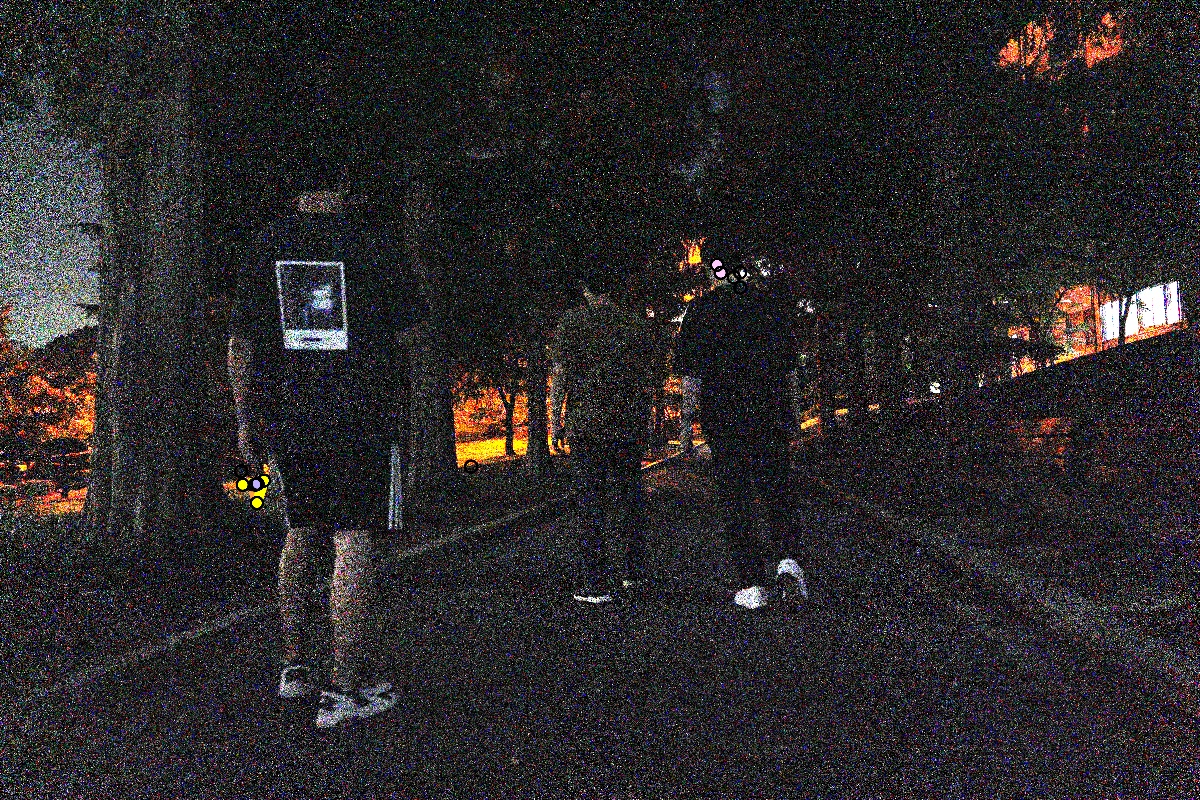}
	\end{minipage}
	\begin{minipage}[c]{0.159\linewidth}
		\includegraphics[width=\linewidth]{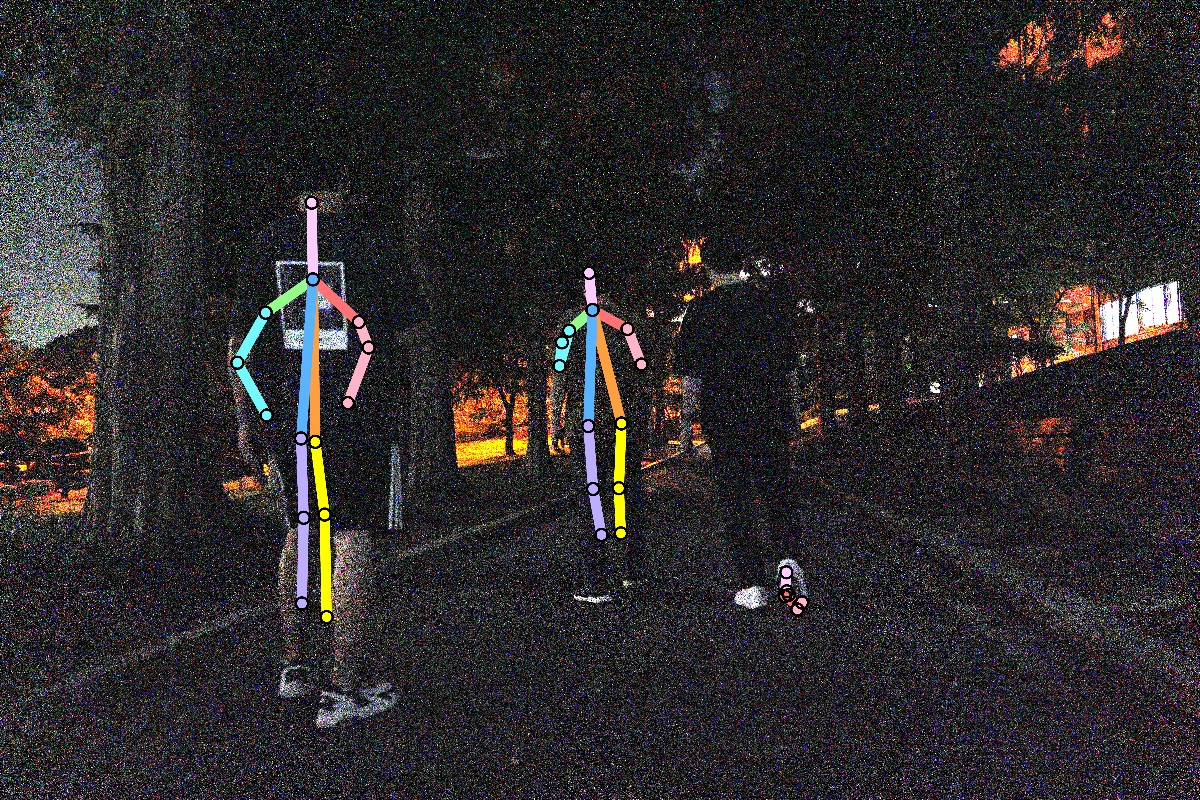}
	\end{minipage}
    \begin{minipage}[c]{0.159\linewidth}
		\includegraphics[width=\linewidth]{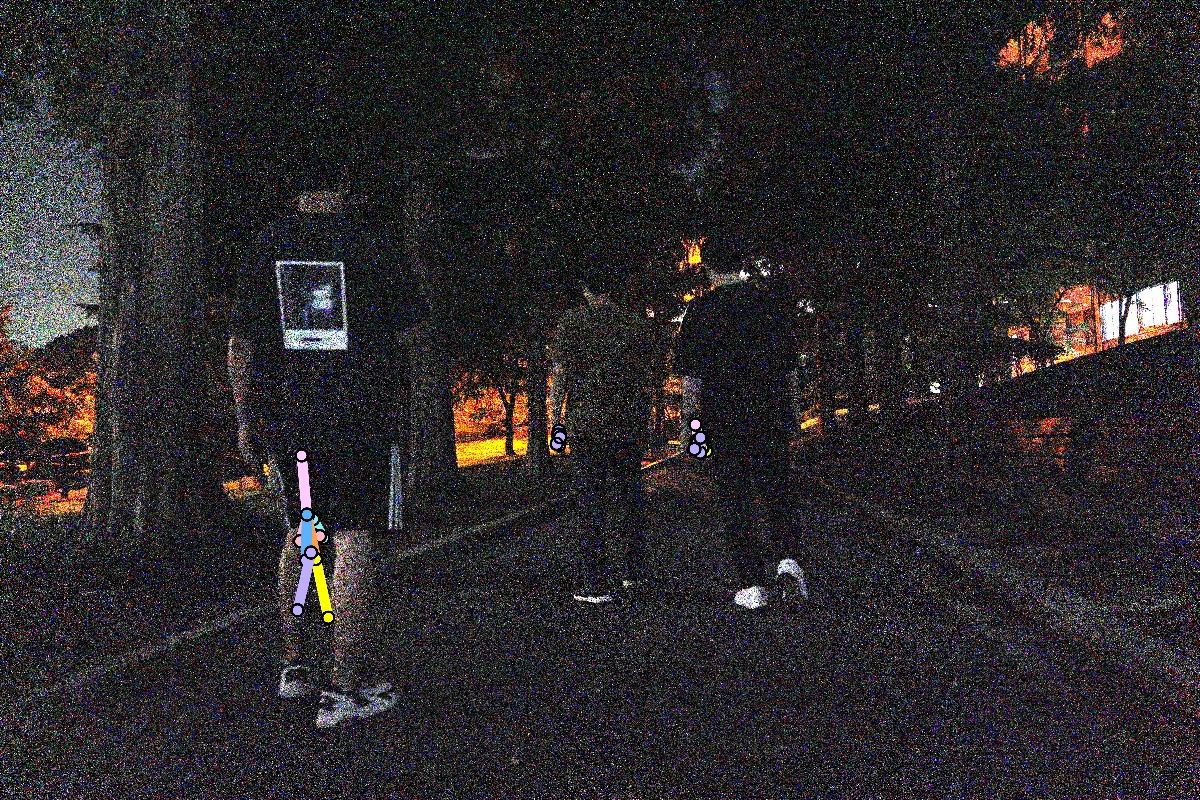}
	\end{minipage}
    \begin{minipage}[c]{0.159\linewidth}
		\includegraphics[width=\linewidth]{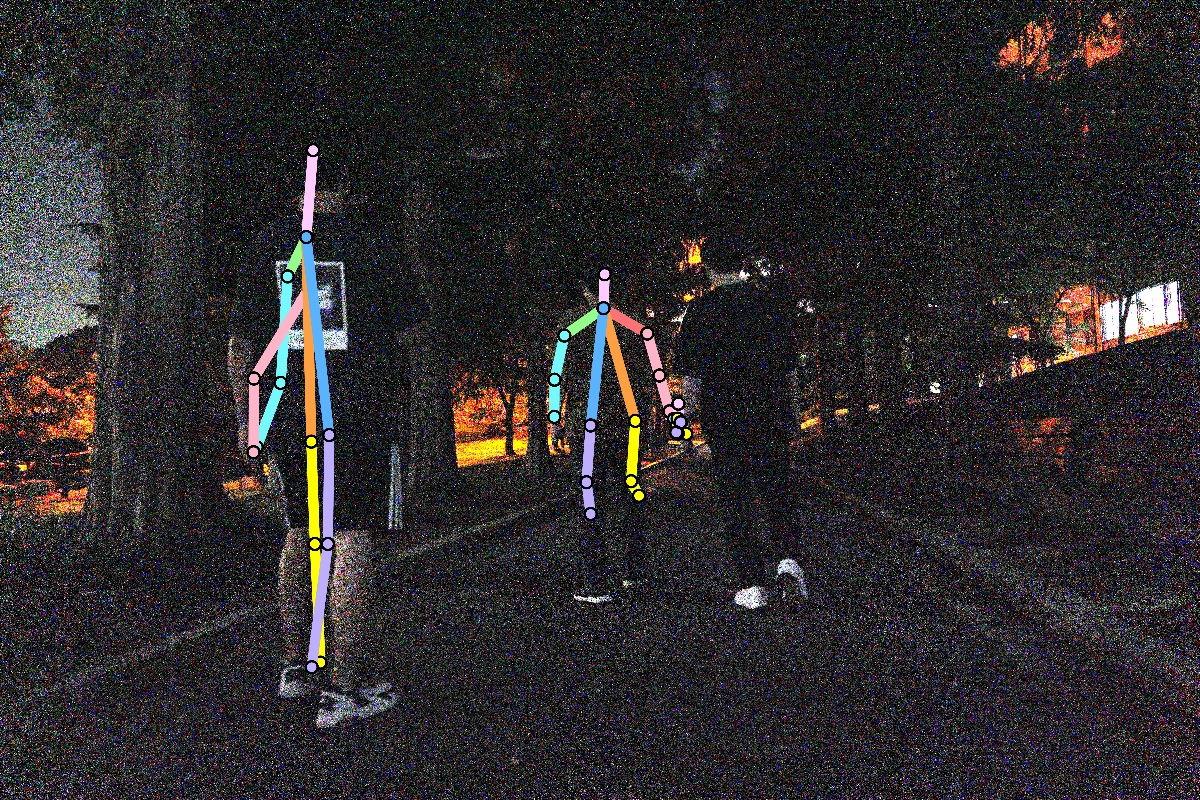}
	\end{minipage}
	\begin{minipage}[c]{0.159\linewidth}
		\includegraphics[width=\linewidth]{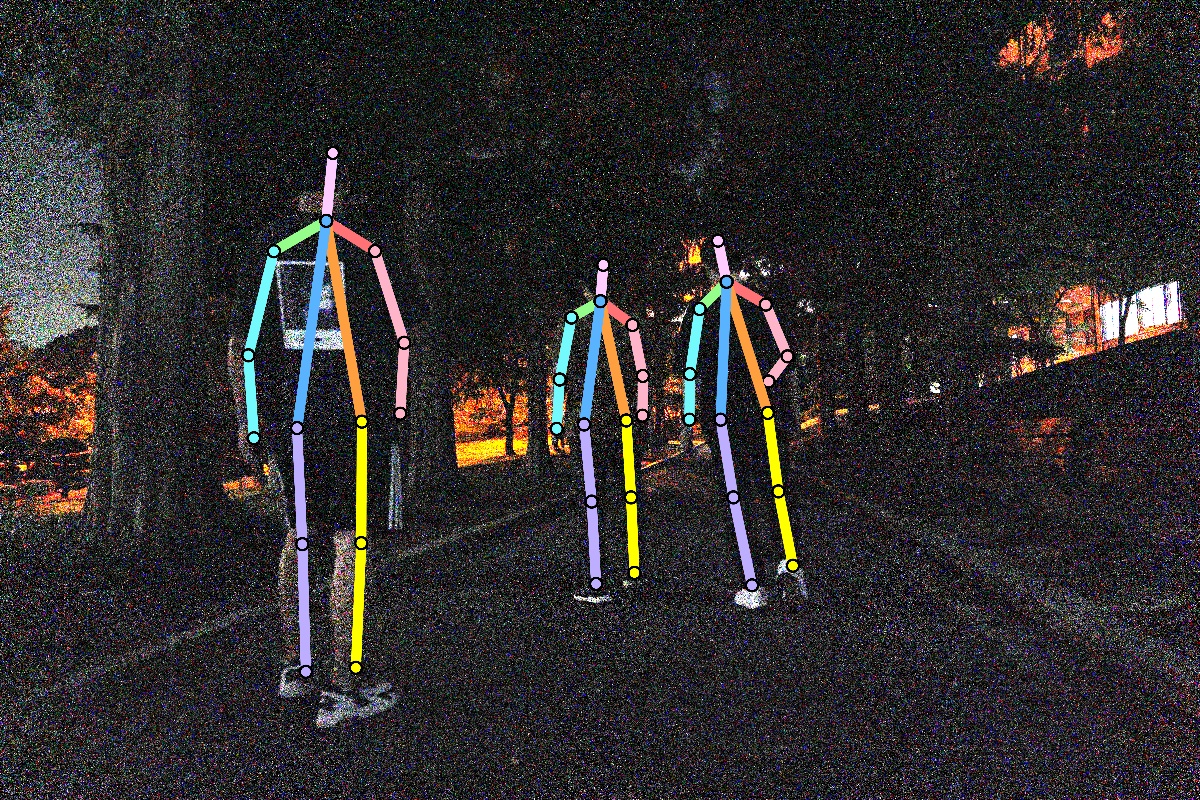}
	\end{minipage}


     \begin{minipage}[c]{0.159\linewidth}
		\includegraphics[width=\linewidth]{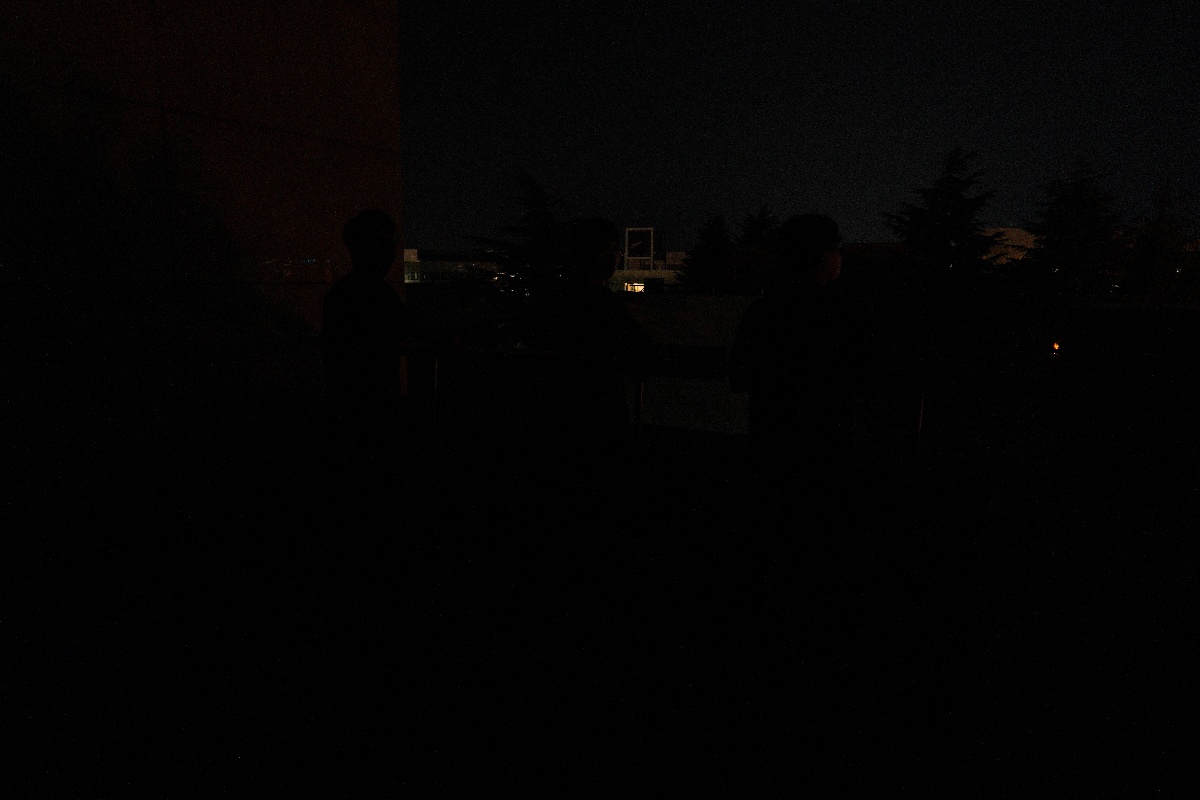}
	\end{minipage}
     \begin{minipage}[c]{0.159\linewidth}
		\includegraphics[width=\linewidth]{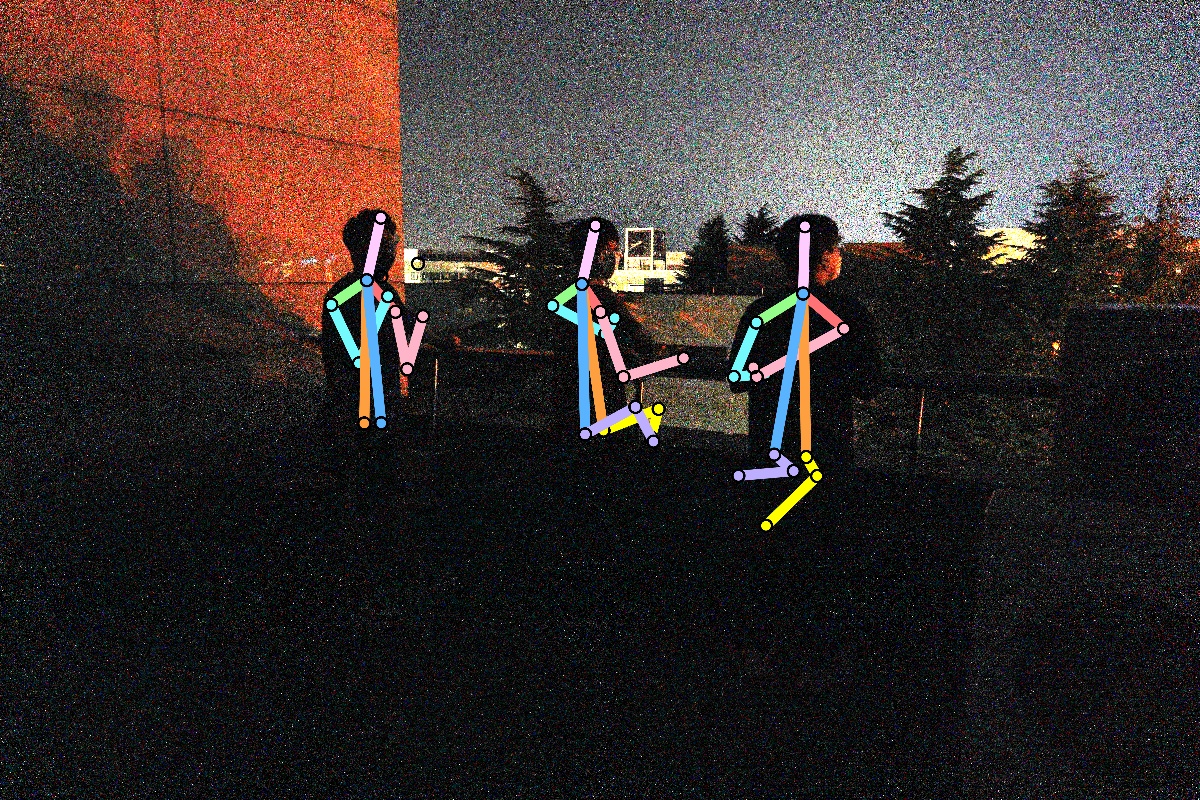}
	\end{minipage}
    \begin{minipage}[c]{0.159\linewidth}
		\includegraphics[width=\linewidth]{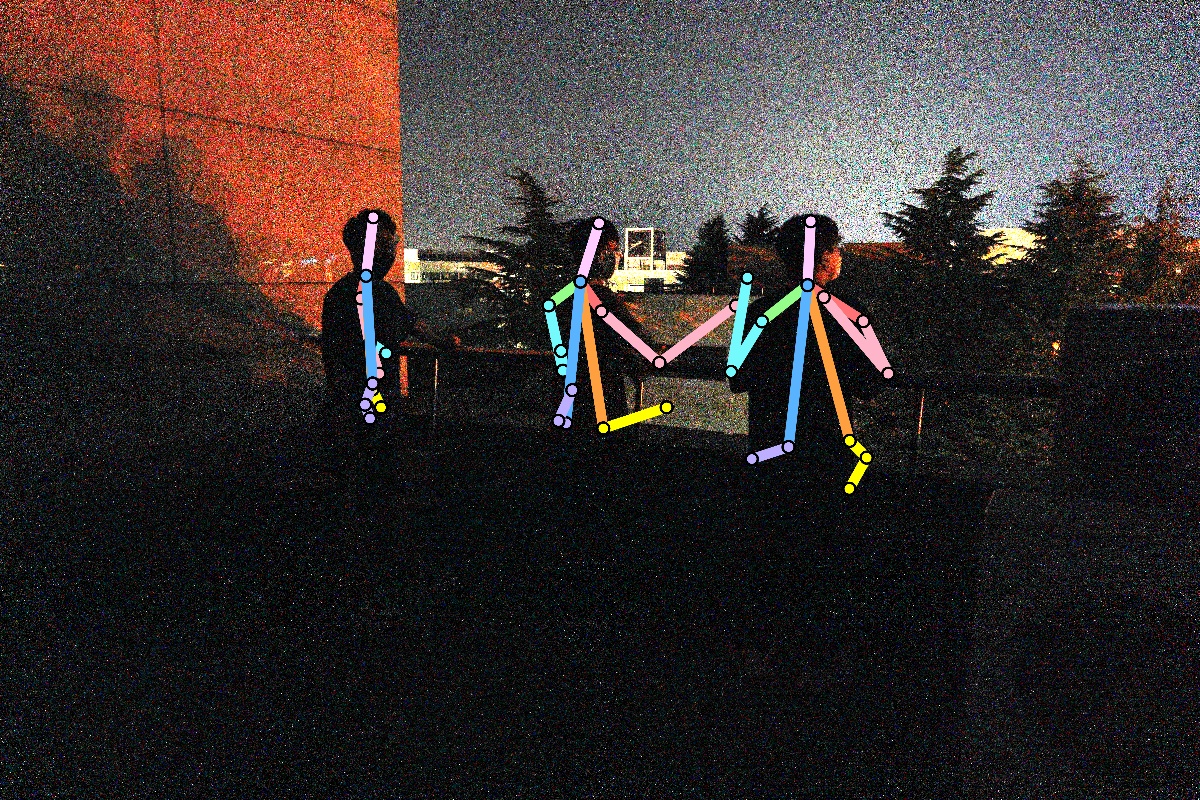}
	\end{minipage}
    \begin{minipage}[c]{0.159\linewidth}
		\includegraphics[width=\linewidth]{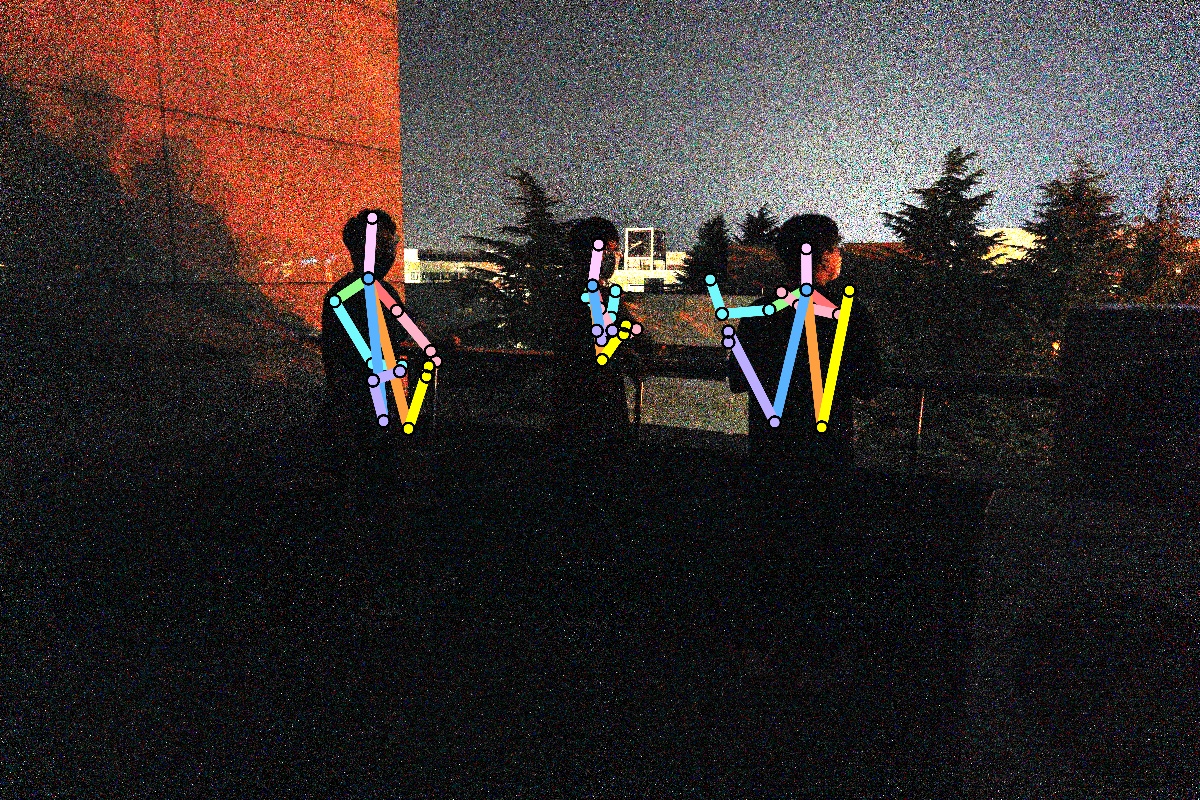}
	\end{minipage}
      \begin{minipage}[c]{0.159\linewidth}
		\includegraphics[width=\linewidth]{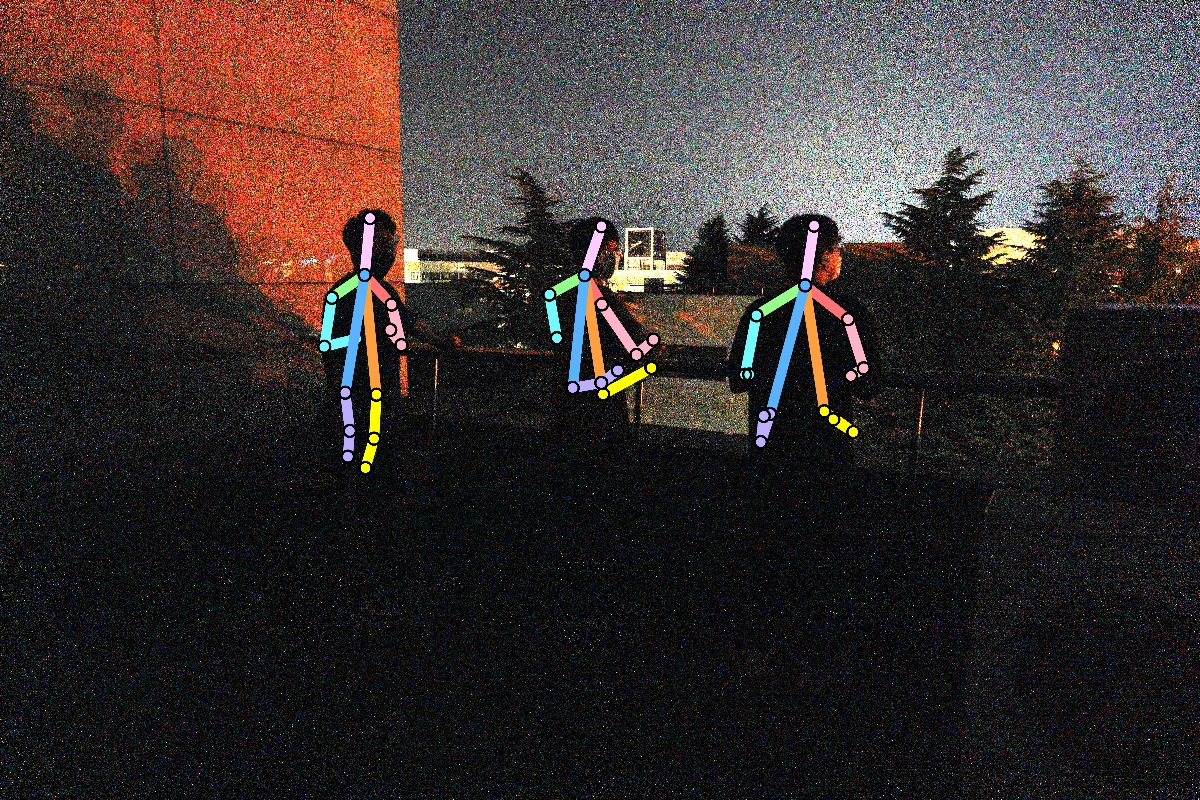}
	\end{minipage}
	\begin{minipage}[c]{0.159\linewidth}
		\includegraphics[width=\linewidth]{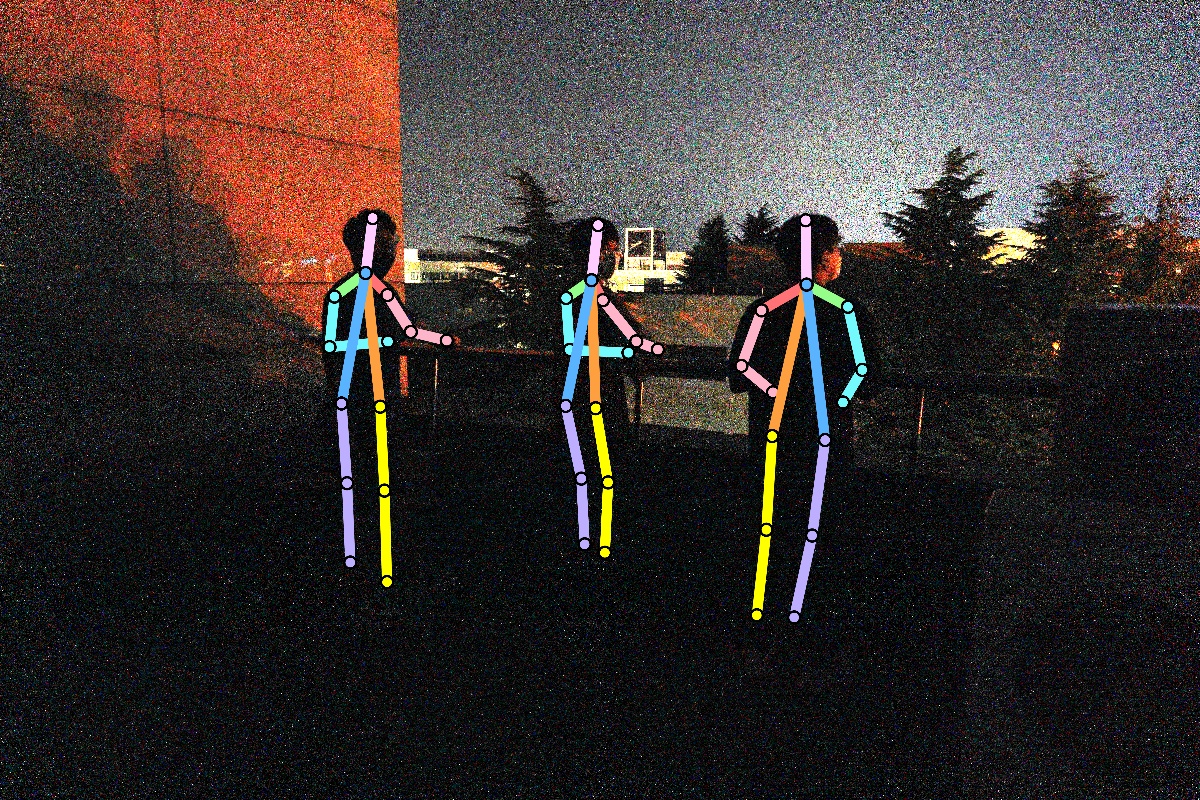}
	\end{minipage}


    \begin{minipage}[c]{0.159\linewidth}
		\includegraphics[width=\linewidth]{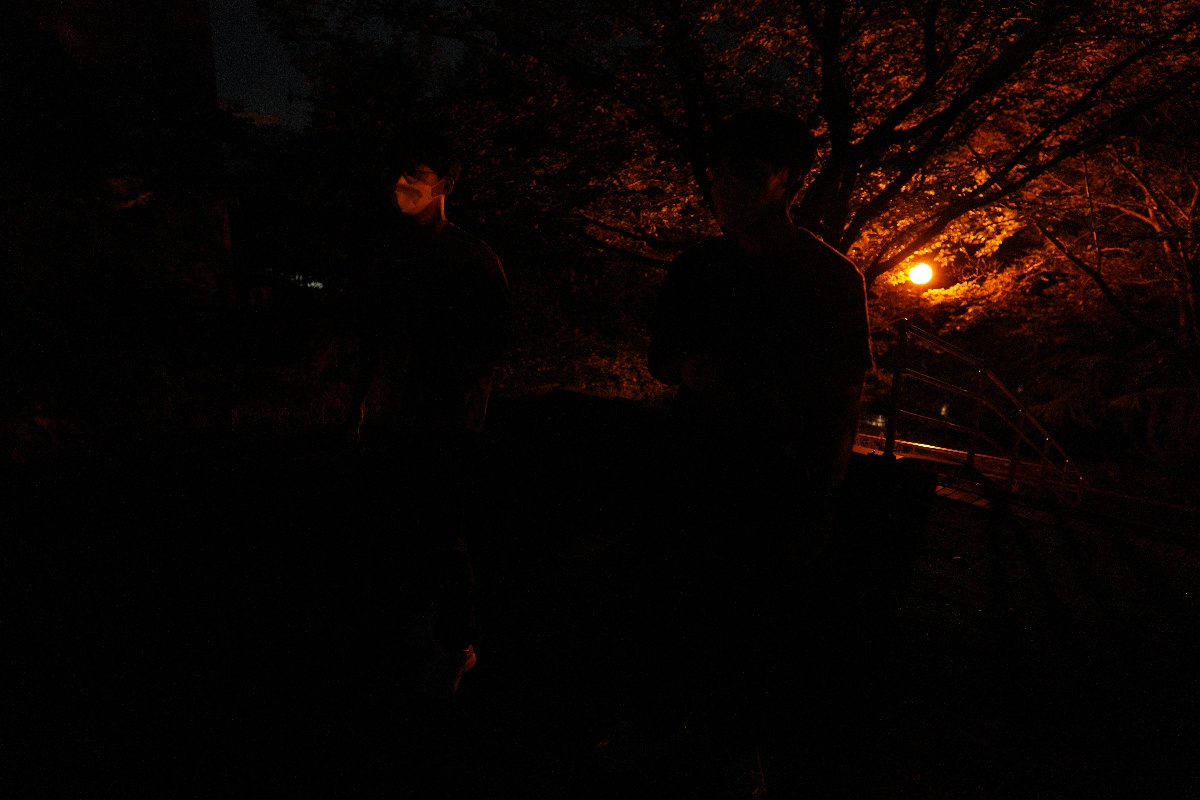}
	\end{minipage}
    \begin{minipage}[c]{0.159\linewidth}
		\includegraphics[width=\linewidth]{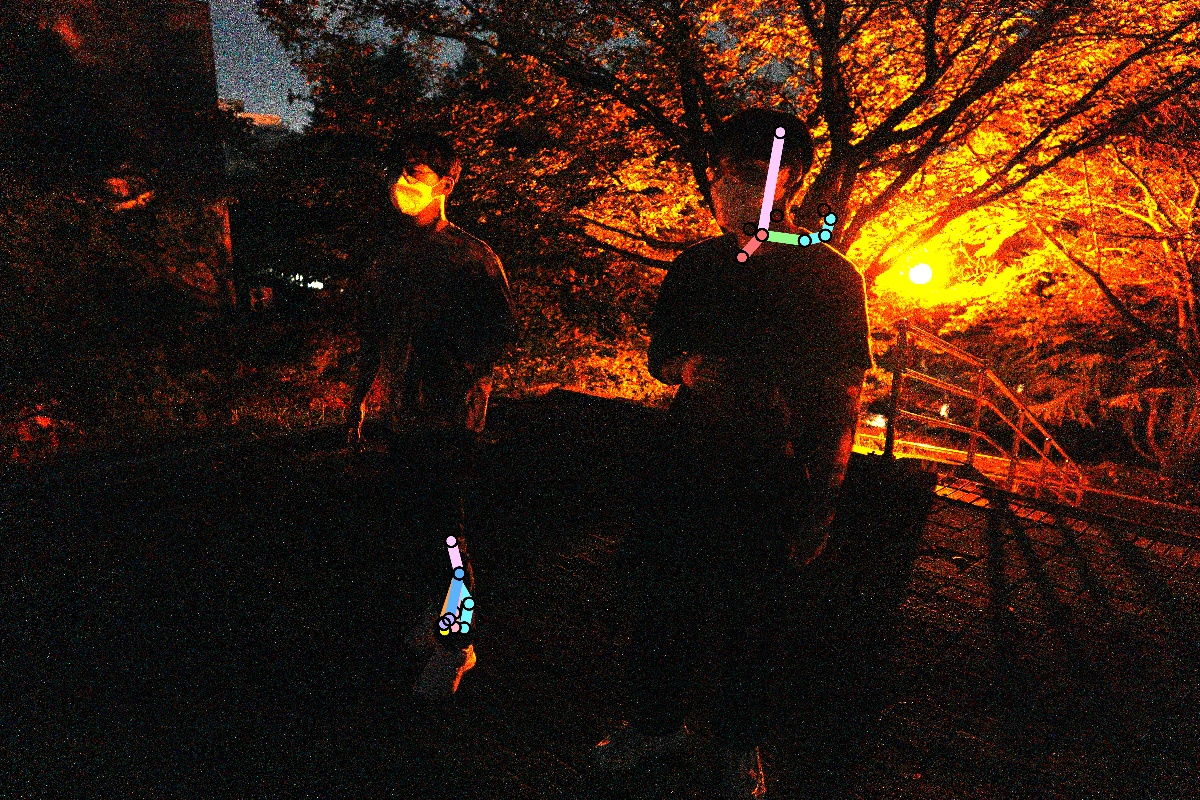}
	\end{minipage}
	\begin{minipage}[c]{0.159\linewidth}
		\includegraphics[width=\linewidth]{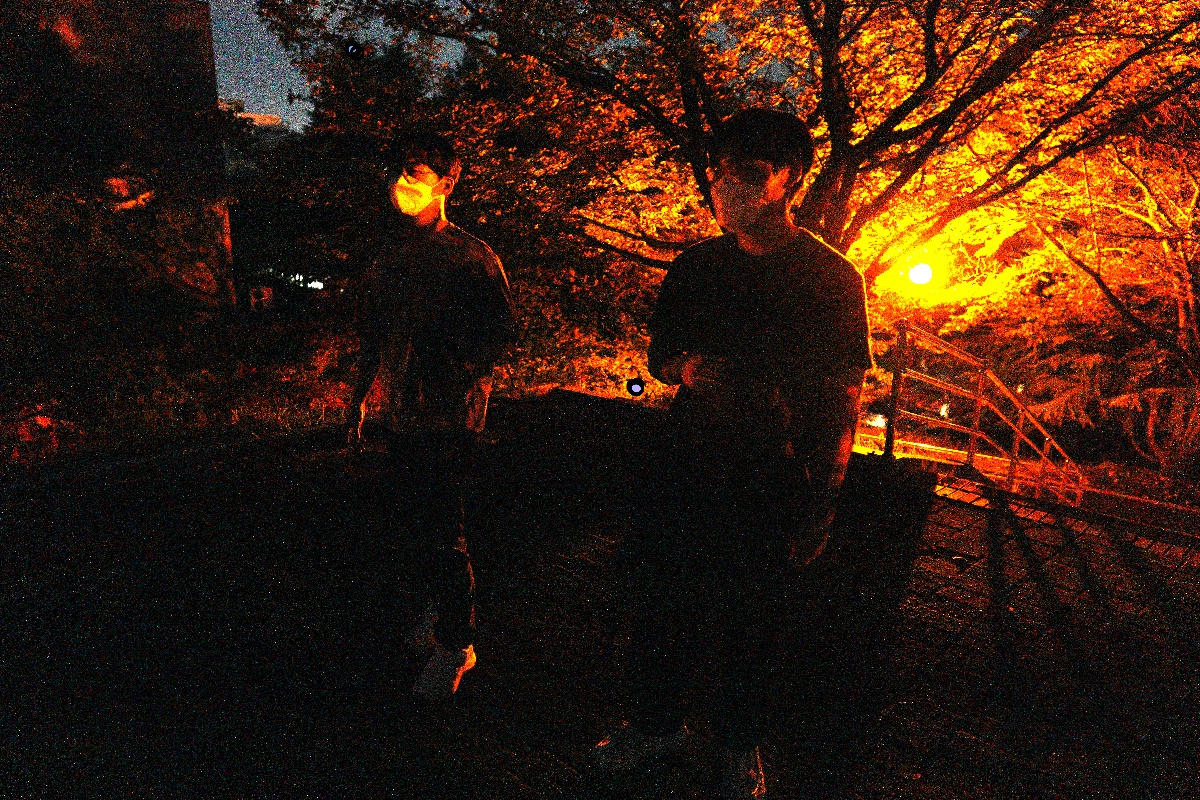}
	\end{minipage}
    \begin{minipage}[c]{0.159\linewidth}
		\includegraphics[width=\linewidth]{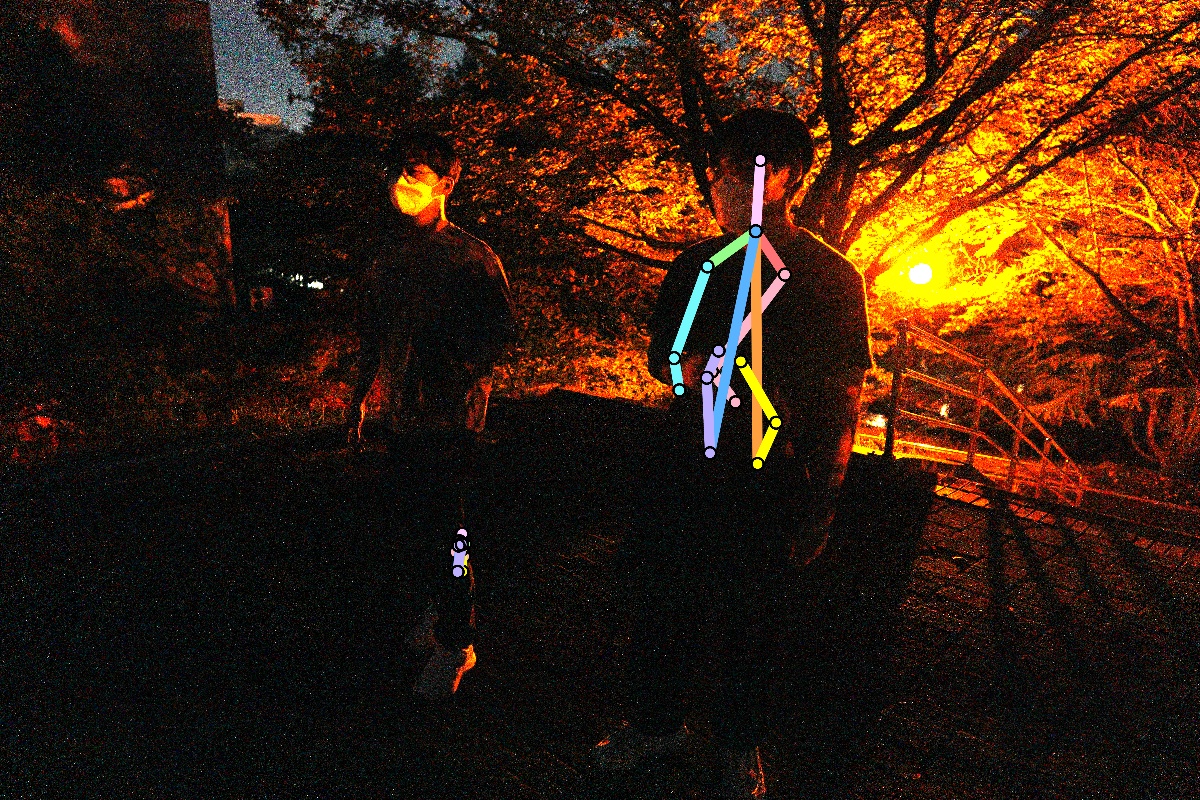}
	\end{minipage}
    \begin{minipage}[c]{0.159\linewidth}
		\includegraphics[width=\linewidth]{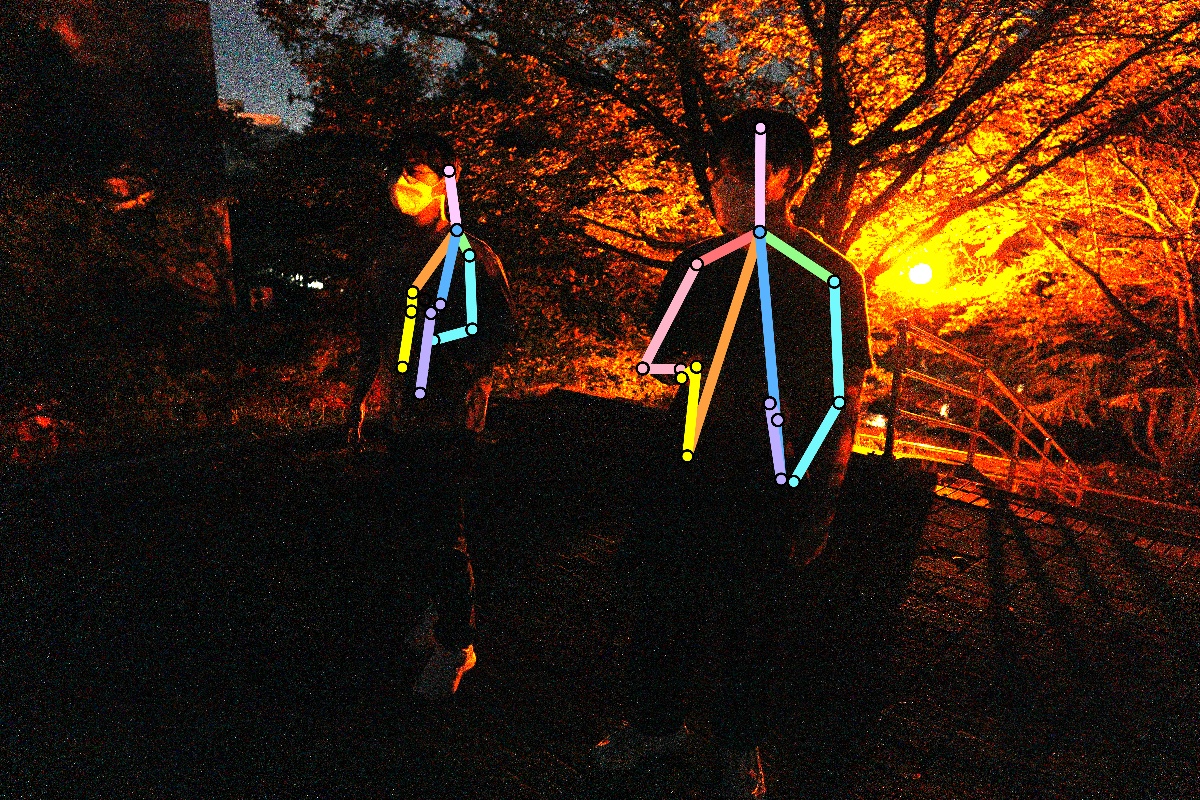}
	\end{minipage}
	\begin{minipage}[c]{0.159\linewidth}
		\includegraphics[width=\linewidth]{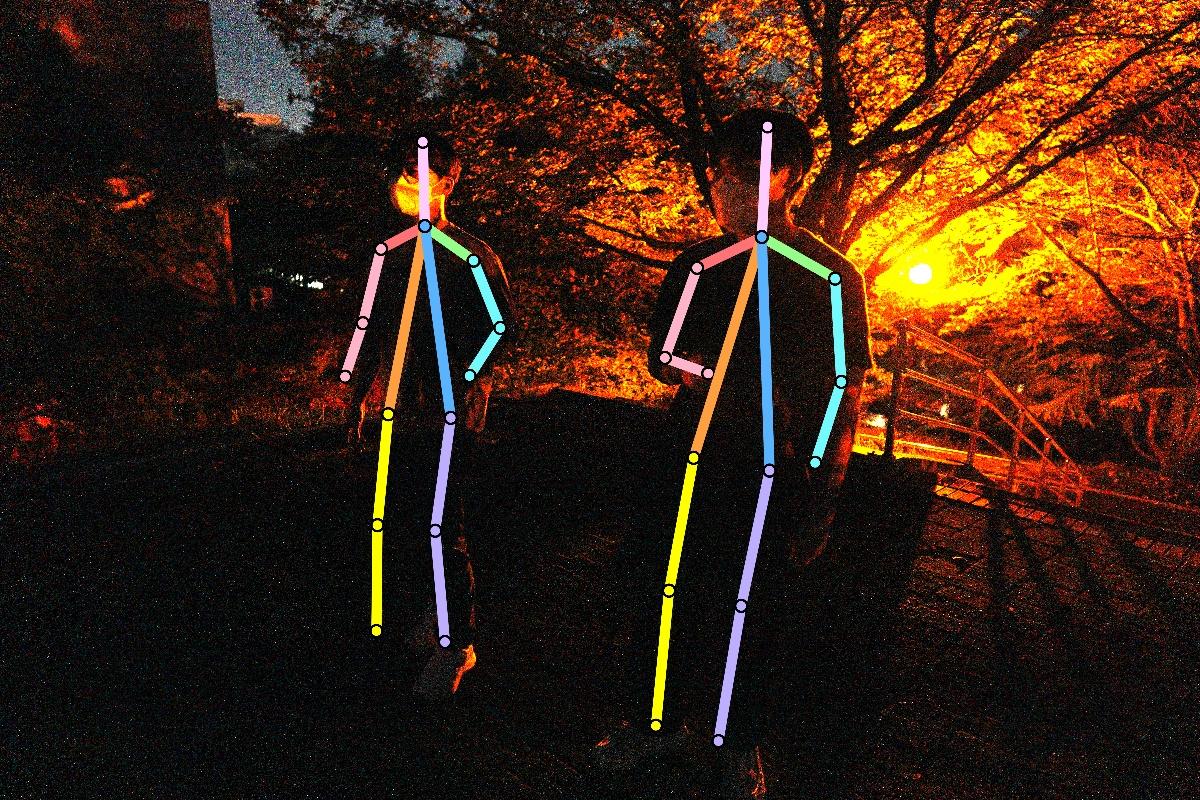}
	\end{minipage}
 
        \begin{minipage}[c]{0.159\linewidth}
		\centerline{\scriptsize{Input Image}}
	\end{minipage}
    \begin{minipage}[c]{0.159\linewidth}
		\centerline{\scriptsize{AdvEnt \cite{AdvEnt}}}
	\end{minipage}
	\begin{minipage}[c]{0.159\linewidth}
		\centerline{\scriptsize{LLFlow \cite{LLFlow}}}
	\end{minipage}	
    \begin{minipage}[c]{0.159\linewidth}
		\centerline{\scriptsize{UDA-HE \cite{kim2022unified}}}
	\end{minipage}
    \begin{minipage}[c]{0.159\linewidth}
		\centerline{\scriptsize{RFormer \cite{cai2023retinexformer}}}
	\end{minipage}	
    \begin{minipage}[c]{0.159\linewidth}
		\centerline{\scriptsize{Ours}}
	\end{minipage}
\captionof{figure}{Qualitative comparison involving two image enhancement methods \cite{LLFlow,cai2023retinexformer}, two domain adaptation methods \cite{AdvEnt,kim2022unified}, and our proposed method on ExLPose-OCN \cite{ExLPose_2023_CVPR}. The human pose results for two image enhancement methods are obtained by initially enhancing the input images from low-light to well-lit and subsequently applying an off-the-shelf human pose estimation method \cite{DEKR2021} to predict the poses. The first column shows the original low-light images, which serve as input for both the baselines and our method. The images in the subsequent columns have been brightened solely for visualization purposes and are not utilized by any of the methods.
}
    
\label{fig:representative_result}

\end{center}

%% file: camere_ready_final_arxiv.bbl
\begin{thebibliography}{10}
\providecommand{\url}[1]{\texttt{#1}}
\providecommand{\urlprefix}{URL }
\providecommand{\doi}[1]{https://doi.org/#1}

\bibitem{cai2023retinexformer}
Cai, Y., Bian, H., Lin, J., Wang, H., Timofte, R., Zhang, Y.: Retinexformer:
  One-stage retinex-based transformer for low-light image enhancement. arXiv
  preprint arXiv:2303.06705  (2023)

\bibitem{cao2019cross}
Cao, J., Tang, H., Fang, H.S., Shen, X., Lu, C., Tai, Y.W.: Cross-domain
  adaptation for animal pose estimation. In: Proceedings of the IEEE/CVF
  international conference on computer vision. pp. 9498--9507 (2019)

\bibitem{cao2019openpose}
Cao, Z., Hidalgo, G., Simon, T., Wei, S.E., Sheikh, Y.: Openpose: realtime
  multi-person 2d pose estimation using part affinity fields. IEEE transactions
  on pattern analysis and machine intelligence  \textbf{43}(1),  172--186
  (2019)

\bibitem{celik2011contextual}
Celik, T., Tjahjadi, T.: Contextual and variational contrast enhancement. IEEE
  Transactions on Image Processing  \textbf{20}(12),  3431--3441 (2011)

\bibitem{chen2018cascaded}
Chen, Y., Wang, Z., Peng, Y., Zhang, Z., Yu, G., Sun, J.: Cascaded pyramid
  network for multi-person pose estimation. In: Proceedings of the IEEE
  conference on computer vision and pattern recognition. pp. 7103--7112 (2018)

\bibitem{cheng2020higherhrnet}
Cheng, B., Xiao, B., Wang, J., Shi, H., Huang, T.S., Zhang, L.: Higherhrnet:
  Scale-aware representation learning for bottom-up human pose estimation. In:
  Proceedings of the IEEE/CVF Conference on Computer Vision and Pattern
  Recognition. pp. 5386--5395 (2020)

\bibitem{cheng2004simple}
Cheng, H.D., Shi, X.: A simple and effective histogram equalization approach to
  image enhancement. Digital signal processing  \textbf{14}(2),  158--170
  (2004)

\bibitem{cheng2023bottom}
Cheng, Y., Ai, Y., Wang, B., Wang, X., Tan, R.T.: Bottom-up 2d pose estimation
  via dual anatomical centers for small-scale persons. Pattern Recognition
  \textbf{139},  109403 (2023)

\bibitem{crescitelli2020poison}
Crescitelli, V., Kosuge, A., Oshima, T.: Poison: Human pose estimation in
  insufficient lighting conditions using sensor fusion. IEEE Transactions on
  Instrumentation and Measurement  \textbf{70}, ~1--8 (2020)

\bibitem{crescitelli2020rgb}
Crescitelli, V., Kosuge, A., Oshima, T.: An rgb/infra-red camera fusion
  approach for multi-person pose estimation in low light environments. In: 2020
  IEEE Sensors Applications Symposium (SAS). pp.~1--6. IEEE (2020)

\bibitem{desmarais2021review}
Desmarais, Y., Mottet, D., Slangen, P., Montesinos, P.: A review of 3d human
  pose estimation algorithms for markerless motion capture. Computer Vision and
  Image Understanding p. 103275 (2021)

\bibitem{dong2019towards}
Dong, H., Liang, X., Shen, X., Wang, B., Lai, H., Zhu, J., Hu, Z., Yin, J.:
  Towards multi-pose guided virtual try-on network. In: Proceedings of the
  IEEE/CVF International Conference on Computer Vision. pp. 9026--9035 (2019)

\bibitem{ganin2016domain}
Ganin, Y., Ustinova, E., Ajakan, H., Germain, P., Larochelle, H., Laviolette,
  F., Marchand, M., Lempitsky, V.: Domain-adversarial training of neural
  networks. The journal of machine learning research  \textbf{17}(1),
  2096--2030 (2016)

\bibitem{DEKR2021}
Geng, Z., Sun, K., Xiao, B., Zhang, Z., Wang, J.: Bottom-up human pose
  estimation via disentangled keypoint regression. In: Proceedings of the
  IEEE/CVF Conference on Computer Vision and Pattern Recognition. pp.
  14676--14686 (2021)

\bibitem{LIME}
Guo, X.: Lime: A method for low-light image enhancement. In: Proceedings of the
  24th ACM international conference on Multimedia. pp. 87--91 (2016)

\bibitem{han2022learning}
Han, Z., Sun, H., Yin, Y.: Learning transferable parameters for unsupervised
  domain adaptation. IEEE Transactions on Image Processing  \textbf{31},
  6424--6439 (2022)

\bibitem{he2017mask}
He, K., Gkioxari, G., Doll{\'a}r, P., Girshick, R.: Mask r-cnn. In: Proceedings
  of the IEEE international conference on computer vision. pp. 2961--2969
  (2017)

\bibitem{hosang2017learning}
Hosang, J., Benenson, R., Schiele, B.: Learning non-maximum suppression. In:
  Proceedings of the IEEE conference on computer vision and pattern
  recognition. pp. 4507--4515 (2017)

\bibitem{huang2023semi}
Huang, L., Li, Y., Tian, H., Yang, Y., Li, X., Deng, W., Ye, J.:
  Semi-supervised 2d human pose estimation driven by position inconsistency
  pseudo label correction module. In: Proceedings of the IEEE/CVF Conference on
  Computer Vision and Pattern Recognition. pp. 693--703 (2023)

\bibitem{huang2017arbitrary}
Huang, X., Belongie, S.: Arbitrary style transfer in real-time with adaptive
  instance normalization. In: Proceedings of the IEEE international conference
  on computer vision. pp. 1501--1510 (2017)

\bibitem{jiang2022avatarposer}
Jiang, J., Streli, P., Qiu, H., Fender, A., Laich, L., Snape, P., Holz, C.:
  Avatarposer: Articulated full-body pose tracking from sparse motion sensing.
  In: European Conference on Computer Vision. pp. 443--460. Springer (2022)

\bibitem{jiang2021regressive}
Jiang, J., Ji, Y., Wang, X., Liu, Y., Wang, J., Long, M.: Regressive domain
  adaptation for unsupervised keypoint detection. In: Proceedings of the
  IEEE/CVF Conference on Computer Vision and Pattern Recognition. pp.
  6780--6789 (2021)

\bibitem{jin2022multibranch}
Jin, R., Zhang, J., Yang, J., Tao, D.: Multibranch adversarial regression for
  domain adaptative hand pose estimation. IEEE Transactions on Circuits and
  Systems for Video Technology  \textbf{32}(9),  6125--6136 (2022)

\bibitem{jin2022unsupervised}
Jin, Y., Yang, W., Tan, R.T.: Unsupervised night image enhancement: When layer
  decomposition meets light-effects suppression. In: European Conference on
  Computer Vision. pp. 404--421. Springer (2022)

\bibitem{2pcnet}
Kennerley, M., Wang, J.G., Veeravalli, B., Tan, R.T.: 2pcnet: Two-phase
  consistency training for day-to-night unsupervised domain adaptive object
  detection. In: Proceedings of the IEEE/CVF Conference on Computer Vision and
  Pattern Recognition. pp. 11484--11493 (2023)

\bibitem{kim2022unified}
Kim, D., Wang, K., Saenko, K., Betke, M., Sclaroff, S.: A unified framework for
  domain adaptive pose estimation. In: European Conference on Computer Vision.
  pp. 603--620. Springer (2022)

\bibitem{kocabas2018multiposenet}
Kocabas, M., Karagoz, S., Akbas, E.: Multiposenet: Fast multi-person pose
  estimation using pose residual network. In: Proceedings of the European
  conference on computer vision (ECCV). pp. 417--433 (2018)

\bibitem{kuhn1955hungarian}
Kuhn, H.W.: The hungarian method for the assignment problem. Naval research
  logistics quarterly  \textbf{2}(1-2),  83--97 (1955)

\bibitem{ExLPose_2023_CVPR}
Lee, S., Rim, J., Jeong, B., Geonu~Kim, B.W., Lee, H., Cho, S., Kwak, S.: Human
  pose estimation in extremely low-light conditions. In: Proceedings of the
  IEEE/CVF Conference on Computer Vision and Pattern Recognition (CVPR) (2023)

\bibitem{li2021synthetic}
Li, C., Lee, G.H.: From synthetic to real: Unsupervised domain adaptation for
  animal pose estimation. In: Proceedings of the IEEE/CVF conference on
  computer vision and pattern recognition. pp. 1482--1491 (2021)

\bibitem{li2019crowdpose}
Li, J., Wang, C., Zhu, H., Mao, Y., Fang, H.S., Lu, C.: Crowdpose: Efficient
  crowded scenes pose estimation and a new benchmark. In: Proceedings of the
  IEEE/CVF Conference on Computer Vision and Pattern Recognition. pp.
  10863--10872 (2019)

\bibitem{li2018structure}
Li, M., Liu, J., Yang, W., Sun, X., Guo, Z.: Structure-revealing low-light
  image enhancement via robust retinex model. IEEE Transactions on Image
  Processing  \textbf{27}(6),  2828--2841 (2018)

\bibitem{lin2022prototype}
Lin, H., Zhang, Y., Qiu, Z., Niu, S., Gan, C., Liu, Y., Tan, M.:
  Prototype-guided continual adaptation for class-incremental unsupervised
  domain adaptation. In: European Conference on Computer Vision. pp. 351--368.
  Springer (2022)

\bibitem{lin2014microsoft}
Lin, T.Y., Maire, M., Belongie, S., Hays, J., Perona, P., Ramanan, D.,
  Doll{\'a}r, P., Zitnick, C.L.: Microsoft coco: Common objects in context. In:
  European conference on computer vision. pp. 740--755. Springer (2014)

\bibitem{moran2020deeplpf}
Moran, S., Marza, P., McDonagh, S., Parisot, S., Slabaugh, G.: Deeplpf: Deep
  local parametric filters for image enhancement. In: Proceedings of the
  IEEE/CVF conference on computer vision and pattern recognition. pp.
  12826--12835 (2020)

\bibitem{mu2020learning}
Mu, J., Qiu, W., Hager, G.D., Yuille, A.L.: Learning from synthetic animals.
  In: Proceedings of the IEEE/CVF Conference on Computer Vision and Pattern
  Recognition. pp. 12386--12395 (2020)

\bibitem{newell2017associative}
Newell, A., Huang, Z., Deng, J.: Associative embedding: End-to-end learning for
  joint detection and grouping. Advances in Neural Information Processing
  Systems  \textbf{30} (2017)

\bibitem{peng2023source}
Peng, Q., Zheng, C., Chen, C.: Source-free domain adaptive human pose
  estimation. In: Proceedings of the IEEE/CVF International Conference on
  Computer Vision. pp. 4826--4836 (2023)

\bibitem{Punnappurath_2022_CVPR}
Punnappurath, A., Abuolaim, A., Abdelhamed, A., Levinshtein, A., Brown, M.S.:
  Day-to-night image synthesis for training nighttime neural isps. In:
  Proceedings of the IEEE/CVF Conference on Computer Vision and Pattern
  Recognition (CVPR). pp. 10769--10778 (June 2022)

\bibitem{rahman2016adaptive}
Rahman, S., Rahman, M.M., Abdullah-Al-Wadud, M., Al-Quaderi, G.D., Shoyaib, M.:
  An adaptive gamma correction for image enhancement. EURASIP Journal on Image
  and Video Processing  \textbf{2016}(1),  1--13 (2016)

\bibitem{raychaudhuri2023prior}
Raychaudhuri, D.S., Ta, C.K., Dutta, A., Lal, R., Roy-Chowdhury, A.K.:
  Prior-guided source-free domain adaptation for human pose estimation. In:
  Proceedings of the IEEE/CVF International Conference on Computer Vision. pp.
  14996--15006 (2023)

\bibitem{sharma2021nighttime}
Sharma, A., Tan, R.T.: Nighttime visibility enhancement by increasing the
  dynamic range and suppression of light effects. In: Proceedings of the
  IEEE/CVF Conference on Computer Vision and Pattern Recognition. pp.
  11977--11986 (2021)

\bibitem{sun2019deep}
Sun, K., Xiao, B., Liu, D., Wang, J.: Deep high-resolution representation
  learning for human pose estimation. In: Proceedings of the IEEE/CVF
  Conference on Computer Vision and Pattern Recognition. pp. 5693--5703 (2019)

\bibitem{tarvainen2017mean}
Tarvainen, A., Valpola, H.: Mean teachers are better role models:
  Weight-averaged consistency targets improve semi-supervised deep learning
  results. Advances in neural information processing systems  \textbf{30}
  (2017)

\bibitem{tian2019directpose}
Tian, Z., Chen, H., Shen, C.: Directpose: Direct end-to-end multi-person pose
  estimation. arXiv preprint arXiv:1911.07451  (2019)

\bibitem{AdvEnt}
Vu, T.H., Jain, H., Bucher, M., Cord, M., P{\'e}rez, P.: Advent: Adversarial
  entropy minimization for domain adaptation in semantic segmentation. In:
  Proceedings of the IEEE/CVF conference on computer vision and pattern
  recognition. pp. 2517--2526 (2019)

\bibitem{wang2022contextual}
Wang, D., Zhang, S.: Contextual instance decoupling for robust multi-person
  pose estimation. In: Proceedings of the IEEE/CVF Conference on Computer
  Vision and Pattern Recognition. pp. 11060--11068 (2022)

\bibitem{wang2021robust}
Wang, D., Zhang, S., Hua, G.: Robust pose estimation in crowded scenes with
  direct pose-level inference. Advances in Neural Information Processing
  Systems  \textbf{34} (2021)

\bibitem{wang2022continual}
Wang, Q., Fink, O., Van~Gool, L., Dai, D.: Continual test-time domain
  adaptation. In: Proceedings of the IEEE/CVF Conference on Computer Vision and
  Pattern Recognition. pp. 7201--7211 (2022)

\bibitem{wang2019underexposed}
Wang, R., Zhang, Q., Fu, C.W., Shen, X., Zheng, W.S., Jia, J.: Underexposed
  photo enhancement using deep illumination estimation. In: Proceedings of the
  IEEE/CVF conference on computer vision and pattern recognition. pp.
  6849--6857 (2019)

\bibitem{wang2013naturalness}
Wang, S., Zheng, J., Hu, H.M., Li, B.: Naturalness preserved enhancement
  algorithm for non-uniform illumination images. IEEE transactions on image
  processing  \textbf{22}(9),  3538--3548 (2013)

\bibitem{LLFlow}
Wang, Y., Wan, R., Yang, W., Li, H., Chau, L.P., Kot, A.: Low-light image
  enhancement with normalizing flow. In: Proceedings of the AAAI conference on
  artificial intelligence. vol.~36, pp. 2604--2612 (2022)

\bibitem{wei2020physics}
Wei, K., Fu, Y., Yang, J., Huang, H.: A physics-based noise formation model for
  extreme low-light raw denoising. In: IEEE Conference on Computer Vision and
  Pattern Recognition (2020)

\bibitem{wei2021physics}
Wei, K., Fu, Y., Zheng, Y., Yang, J.: Physics-based noise modeling for extreme
  low-light photography. IEEE Transactions on Pattern Analysis and Machine
  Intelligence  \textbf{44}(11),  8520--8537 (2021)

\bibitem{weng2019photo}
Weng, C.Y., Curless, B., Kemelmacher-Shlizerman, I.: Photo wake-up: 3d
  character animation from a single photo. In: Proceedings of the IEEE/CVF
  Conference on Computer Vision and Pattern Recognition. pp. 5908--5917 (2019)

\bibitem{xiao2018simple}
Xiao, B., Wu, H., Wei, Y.: Simple baselines for human pose estimation and
  tracking. In: Proceedings of the European conference on computer vision
  (ECCV). pp. 466--481 (2018)

\bibitem{xie2021empirical}
Xie, R., Wang, C., Zeng, W., Wang, Y.: An empirical study of the collapsing
  problem in semi-supervised 2d human pose estimation. In: Proceedings of the
  IEEE/CVF International Conference on Computer Vision. pp. 11240--11249 (2021)

\bibitem{LOGOCAP}
Xue, N., Wu, T., Xia, G.S., Zhang, L.: Learning local-global contextual
  adaptation for multi-person pose estimation. In: IEEE Conference on Computer
  Vision and Pattern Recognition (CVPR) (2022)

\bibitem{yan2018spatial}
Yan, S., Xiong, Y., Lin, D.: Spatial temporal graph convolutional networks for
  skeleton-based action recognition. In: Thirty-second AAAI conference on
  artificial intelligence (2018)

\bibitem{zhang2019pose2seg}
Zhang, S.H., Li, R., Dong, X., Rosin, P., Cai, Z., Han, X., Yang, D., Huang,
  H., Hu, S.M.: Pose2seg: Detection free human instance segmentation. In:
  Proceedings of the IEEE/CVF Conference on Computer Vision and Pattern
  Recognition. pp. 889--898 (2019)

\end{thebibliography}
